\documentclass[10pt,twocolumn,letterpaper]{article}

\usepackage[pagenumbers]{cvpr} 

\usepackage{booktabs}
\usepackage[acronym]{glossaries}
\usepackage{enumitem}

\usepackage{amsmath}
\usepackage{amssymb}
\usepackage{amsthm}
\usepackage{mathtools}
\usepackage{bm}

\usepackage{algorithm}
\usepackage{algpseudocode}

\usepackage{graphicx}
\usepackage{epstopdf}
\usepackage{caption}
\usepackage{subcaption}

\usepackage[pagebackref,breaklinks,colorlinks]{hyperref}

\newacronym{DNN}{DNN}{deep neural network}
\newacronym{SGD}{SGD}{stochastic gradient descent}
\newacronym{SNN}{SNN}{spiking neural network}
\newacronym{TinyML}{TinyML}{tiny machine learning}
\newacronym{QNN}{QNN}{quantised neural network}
\newacronym{STE}{STE}{straight-through estimator}
\newacronym{ANA}{ANA}{additive noise annealing}
\newglossaryentry{DoF}
{
    name={DoF},
    description={degree of freedom},
    first={\glsentrydesc{DoF} (\glsentrytext{DoF})},
    plural={DoFs},
    firstplural={degrees of freedom (\glsentryplural{DoF})}
}
\newacronym{SVHN}{SVHN}{Street View House Numbers}
\newacronym{GSC}{GSC}{Google Speech Commands}

\newcommand{\lam}[1]{{#1}_{\lambda}}

\newcommand{\E}{\mathbb{E}}             
\newcommand{\Z}{\mathbb{Z}}             
\newcommand{\R}{\mathbb{R}}             
\renewcommand{\l}{\ell}                 
\newcommand{\bl}{\bar{\ell}}            
\newcommand{\ph}{\varphi}               
\newcommand{\x}{\mathbf{x}}             
\newcommand{\m}{\mathbf{m}}             
\renewcommand{\hm}{\mathbf{\hat{m}}}    
\newcommand{\hM}{\widehat{M}}           
\newcommand{\w}{\mathbf{w}}             
\newcommand{\W}{\mathbf{W}}             
\renewcommand{\b}{\mathbf{b}}           
\newcommand{\s}{\mathbf{s}}             
\newcommand{\eps}{\varepsilon}
\newcommand{\g}{\mathbf{g}}             

\DeclareMathOperator{\clip}{clip}
\DeclareMathOperator{\argmax}{\arg\max}

\newtheorem{theorem}{Theorem}
\newtheorem{lemma}[theorem]{Lemma}
\newtheorem{proposition}[theorem]{Proposition}

\def\cvprPaperID{11171} 
\def\confName{CVPR}
\def\confYear{2022}

\begin{document}

\title{Training Quantised Neural Networks with STE Variants: \\ the Additive Noise Annealing Algorithm}

\author{%
Matteo Spallanzani$^{1}$\thanks{Corresponding author: \texttt{spmatteo@ethz.ch}} \quad
Gian Paolo Leonardi$^{2}$ \quad
Luca Benini$^{1,3}$\\
$^{1}$\small{Departement Informationstechnologie und Elektrotechnik, ETH Z\"{u}rich, Switzerland}\\
$^{2}$\small{Dipartimento di Matematica, Universit\`{a} di Trento, Italy}\\
$^{3}$\small{Dipartimento di Ingegneria Elettrica e dell'Informazione, Universit\`{a} di Bologna, Italy}
}

\maketitle

\begin{abstract}
    Training \glspl{QNN} is a non-differentiable optimisation problem since weights and features are output by piecewise constant functions.
    The standard solution is to apply the \gls{STE}, using different functions during the inference and gradient computation steps.
    Several \gls{STE} variants have been proposed in the literature aiming to maximise the task accuracy of the trained network.
    In this paper, we analyse \gls{STE} variants and study their impact on \gls{QNN} training.
    We first observe that most such variants can be modelled as stochastic regularisations of stair functions; although this intuitive interpretation is not new, our rigorous discussion generalises to further variants.
    Then, we analyse \glspl{QNN} mixing different regularisations, finding that some suitably synchronised smoothing of each layer map is required to guarantee pointwise compositional convergence to the target discontinuous function.
    Based on these theoretical insights, we propose \gls{ANA}, a new algorithm to train \glspl{QNN} encompassing standard \gls{STE} and its variants as special cases.
    When testing \gls{ANA} on the CIFAR-10 image classification benchmark, we find that the major impact on task accuracy is not due to the qualitative shape of the regularisations but to the proper synchronisation of the different \gls{STE} variants used in a network, in accordance with the theoretical results.
\end{abstract}

\section{Introduction}
\label{sec:intro}

Deep learning has rapidly advanced during the last decade over a wide range of domains, from computer vision to natural language processing \cite{He2016, Brown2020}.
However, making it pervasive requires deploying \glspl{DNN} on embedded or edge devices, i.e., on computing systems with limited storage, memory, and processing capabilities.
These constraints are at odds with the typical requirements of \glspl{DNN}, which need millions or even billions of parameters and operations to deliver their performance.

Research in \gls{TinyML} has made considerable steps towards enabling the deployment of \glspl{DNN} on resource-constrained devices.
A first class of techniques aims at making \glspl{DNN} more efficient in terms of accuracy-per-parameter or accuracy-per-operation \cite{Tan2019, Tan2020}.
We refer to these techniques as \textit{topological optimisations} since they revolve around achieving higher model efficiency by changing the structure of \glspl{DNN}.
A second class of approaches has instead focussed on deriving models that leverage the properties of the target deployment platform.
These \textit{hardware-related optimisations} include hardware-friendly activation functions \cite{Nair2010}, weight clustering and weight tensor decomposition \cite{Han2016, Kolda2009, Zhang2016}, and \glspl{QNN} \cite{Hubara2018}.

\Glspl{QNN} use reduced-precision integer operands to meet the storage requirements and exploit the optimised support for integer arithmetic of embedded and edge platforms.
With respect to their floating-point counterparts, \glspl{QNN} typically introduce drops in task accuracy \cite{Jacob2018}.
Several strategies have been proposed to counteract this shortcoming, ranging from changes to the target \gls{QNN}'s topology \cite{Lin2017, Zhuang2019}, through mixing different data representations and precisions inside the same network \cite{Liu2018, vanBaalen2020}, to learning the shape of the stair functions used \cite{Choi2019, Esser2020, Jain2020}.
However, understanding how to propagate gradients through the discontinuous functions that model quantised operands remains a critical problem in \gls{QNN} training.

\paragraph{Background \& related work}
The classical derivative of a piecewise constant function is zero at all continuity points, while its distributional derivative is a linear combination of Dirac's deltas.
Therefore, one can not directly apply the backpropagation algorithm to train a \gls{QNN}.
The standard solution to this problem is applying the so-called \gls{STE} to all the discontinuous functions in the target \gls{QNN} \cite{Bengio2013, Hubara2016}.
Applying \gls{STE} amounts to using two different functions during the forward and backward steps of the learning iteration, with the second one being differentiable.
The choice of the replacement function is not unique.
Previous research suggested that the alternative gradient computed using \gls{STE} replacements is a descent direction for the so-called population loss, and that choosing proper backward functions is necessary to ensure convergence to a local minimum of the loss landscape \cite{Li2017, Yin2019}.

Several \gls{STE} variants have been proposed in the literature aiming at maximising the task accuracy of the trained \gls{QNN} \cite{Deng2018, Liu2018, Wu2018}.
It is worth noting that the problem of propagating gradients through non-differentiabilities is also relevant for \glspl{SNN} \cite{Wu2018, Severa2019}.
Indeed, research in \gls{SNN} training algorithms has proposed surrogate gradients similar to \gls{STE}.
In particular, the Whetstone method offers a solution to train an \gls{SNN} from a \gls{DNN} \cite{Severa2019}: the method gradually transforms a \gls{DNN} into an \gls{SNN} by annealing the \gls{DNN}'s activation functions to the Heaviside step function during training.
The annealing follows a heuristic schedule where the closer an activation is to the input, the sooner it is annealed.
A similar soft-to-hard annealing has also been proposed to compress images into binary representations using autoencoders \cite{Agustsson2017}.
These dynamic \gls{STE} variants add to the static variants proposed for \glspl{QNN}.
Is it possible to derive a unified description of static and dynamic \gls{STE} variants?
And what is their impact on \gls{QNN} training?

\paragraph{Main contributions}
In this paper, we propose a theoretical framework to describe \gls{STE} variants in a unified way.
Specifically, we provide the following contributions to the field of \gls{QNN} research:
\begin{itemize}
    \item we observe that the backward functions associated with several \gls{STE} variants proposed in the literature can be represented as the expected value of quantisers processing noisy inputs; this interpretation originates numerous families of \gls{STE} variants;
    \item we analyse the problem of applying dynamic variants of \gls{STE} to \glspl{QNN}, introducing the novel concept of \textit{compositional convergence};
    \item we introduce \gls{ANA}, a new algorithm to train \glspl{QNN} encompassing standard \gls{STE} and its variants (both static and dynamic ones) as special cases;
    \item when applying \gls{ANA} to the CIFAR-10 image classification benchmark, we observe that the impact of the qualitative shape of the \gls{STE} backward function on the final accuracy is at best minor; instead, we observe that the proper synchronisation of the regularisations in a \gls{QNN} using dynamic \gls{STE} variants is essential to guarantee convergence; the code to reproduce our experiments is available on GitHub\footnote{\url{https://github.com/pulp-platform/quantlab/tree/ANA}}.
\end{itemize}

The paper is organised as follows: in Section~\ref{sec:formulation}, we introduce the required terminology and prove the theoretical results; in Section~\ref{sec:ana}, we describe \gls{ANA}; in Section~\ref{sec:experiments}, we describe our experimental design and discuss its outcomes; finally, in Section~\ref{sec:conclusions}, we summarise our findings.

\section{Analysing \gls{STE} variants}
\label{sec:formulation}

\subsection{Quantisers and \gls{STE}}

Given an integer $K \geq 2$, a set $Q \coloneqq \{ q_{0} < \dots < q_{K-1} \} \subset \R$ of \textbf{quantisation levels} and a set $\Theta \coloneqq \{ \theta_{1} < \dots < \theta_{K-1} \} \subset \R$ of \textbf{thresholds}, we define a \textbf{$K$-quantiser} to be the stair function
\begin{equation}\label{eq:quantiser}
\begin{split}
    \sigma_{\Theta, Q} \,:\,
    \R &\to     Q \\
    x  &\mapsto q_{0} + \sum_{k=1}^{K-1} (q_{k} - q_{k-1}) H^{+}_{\theta_{k}}(x) \,.
\end{split}
\end{equation}
Here,
\begin{equation}\label{eq:heaviside}
\begin{split}
    H^{+}_{\theta} \,:\,
    \R &\to \{ 0, 1 \} \\
    x  &\mapsto
    \begin{cases}
        0, \text{if } x < \theta \,, \\
        1, \text{if } \theta \leq x \,,
    \end{cases}
\end{split}
\end{equation}
is the (parametric) Heaviside function.
Note that $H^{+}_{\theta}$ is itself a $2$-quantiser.
For convenience, we define $K$-quantiser's \textbf{bins} to be the counterimages of the quantisation levels: $I_{k} \coloneqq \sigma_{\Theta, Q}^{-1}(\{ q_{k} \})$.

In practical applications, $K$ is set to be equal to $2^{B}$ for some integer \textbf{precision} $B \geq 1$, and there exist an \textbf{offset} $z \in \Z$ and a \textbf{quantum} $\epsilon \in \R^{+}$ such that $q_{0} = z \epsilon$ and $\theta_{k} = q_{k} = (z + k) \epsilon$ for $k = 1, \dots, K - 1$.
This simplification allows rewriting \eqref{eq:quantiser} in terms of hardware-efficient flooring and clipping operations, yielding a \textbf{linear $B$-bit quantiser}: $\sigma_{\Theta, Q}(x) = \sigma_{z, \epsilon}(x) = \epsilon \clip(\lfloor x / \epsilon \rfloor, z, z + K - 1)$.
$z$ is usually chosen to be $0$ (unsigned $B$-bit linear quantiser) or $-2^{B-1}$ (signed $B$-bit linear quantiser).

When the exact value of $K$ (respectively, $B$) is not relevant or can be inferred from the context, we will simply use the terms \textbf{quantiser} (respectively, \textbf{linear quantiser}).
When clarity of exposition is not impacted, we will also drop the subscripts to ease readability.

Quantisers are piecewise constant functions: their classical derivative does not exist at the thresholds and is zero in the interior of the bins.
This lack of differentiability is disruptive for backpropagation, and would theoretically prevent gradient-based training of \glspl{QNN}.

The \gls{STE} can be regarded as a technique to make a quantiser \eqref{eq:quantiser} differentiable by replacing it with a differentiable or almost everywhere differentiable function $\tilde{\sigma}$ before computing the derivative.
We name $\sigma$ the \textbf{\gls{STE} target} and $\tilde{\sigma}$ the \textbf{\gls{STE} regularisation}.
Example replacement functions for the Heaviside $H^{+}_{0}$ include the hard sigmoid $\tilde{\sigma}(x) \coloneqq \max\{ 0, \min\{ x + 1/2, 1 \} \}$, the clipped ReLU $\tilde{\sigma}(x) \coloneqq \max\{ 0, \min\{ x, 1 \} \}$, and the ReLU $\tilde{\sigma}(x) = \max\{ 0, x \}$.

\subsection{Unifying \gls{STE} variants}

Consider a discontinuous function $\sigma : \R \to \R$.
We say that a function $\lam{\sigma} : \R \to \R$ is a corresponding \textbf{parametric regularisation} if $\lam{\sigma} \in C^{1}(\R)$ (or Lipschitz, and thus almost everywhere differentiable) for each $\lambda \in \R^{+}$, and $\lim_{\lambda \to 0} \lam{\sigma}(x) = \sigma(x), \forall x \in \R$; $\lambda \in \R^{+}$ is a parameter controlling the degree of regularisation.

It is elementary to show that the expectation operator acts as a convolution, transforming a discontinuous function into one that is differentiable, either in the classical or distributional sense.
In what follows, we will overload $\mu$ to denote both an absolutely continuous probability measure on $\R$ and the corresponding probability density function.
\begin{proposition}\label{th:regularisation}
Let $\sigma \,:\, \R \to Q$ be a $K$-quantiser.
Let $\nu$ be a real random variable with probability density $\mu$.
For any value $\nu' \in \R$, we define the function $\sigma_{\nu'}(x) \coloneqq \sigma(x - \nu')$.
Then:
\begin{itemize}
    \item[(i)] $\E_{\mu}[\sigma_{\nu}(x)] = (\mu \ast \sigma)(x), \,\forall\, x \in \R$; we therefore define $\E_{\mu}[\sigma_{\nu}] \coloneqq \mu \ast \sigma$;
    \item[(ii)] if $\mu \in W^{1,1}(\R)$ then $\E_{\mu}[\sigma_{\nu}]$ is differentiable, its derivative is bounded, continuous, and satisfies $\frac{d}{dx}\E_{\mu}[\sigma_{\nu}] = D\mu \ast \sigma$;
    \item[(iii)] if $\mu \in BV(\R)$ then $D\E_{\mu}[\sigma_{\nu}] = D\mu \ast \sigma$ almost everywhere and it is bounded by $\|\sigma\|_{\infty} \, |D\mu|(\R)$.
\end{itemize}
\end{proposition}
In other words, the proposition states that we can regularise a discontinuous function $\sigma$ by convolving it with a probability density $\mu$ satisfying either $\mu \in W^{1,1}(\R)$ (e.g., the triangular, normal, and logistic distributions) or the weaker $\mu \in BV(\R)$ (e.g., the uniform distribution on a compact interval).
Note that the noise density is an even function ($\mu(\nu) = \mu(-\nu)$) for common zero-mean distributions such as the uniform, triangular, normal, and logistic.

We can link the concept of regularised functions with the stochastic setting by considering a parametric density $\lam{\mu}$.
As an example, consider the uniform distribution whose mean $\alpha(\lambda)$ and standard deviation $\beta(\lambda)$ depend on $\lambda$ in such a way that $\alpha(\lambda)$ and $\beta(\lambda)$ go to zero when $\lambda \to 0$.
This distribution has density $\lam{\mu}(\nu) = \chi_{[a(\lambda), b(\lambda)]}(\nu) / (b(\lambda) - a(\lambda))$, where $a(\lambda) \coloneqq \alpha(\lambda) - \sqrt{3} \beta(\lambda)$ and $b(\lambda) \coloneqq \alpha(\lambda) + \sqrt{3} \beta(\lambda)$.
Supposing that the target quantiser is $\sigma = H^{+}_{0}$, we have
\begin{equation}\label{eq:ste_uniform}
\begin{split}
    \lam{\sigma}(x) &\coloneqq (\lam{\mu} \ast \sigma)(x) \\
    &=
    \begin{cases}
        0, \,\text{if } x < a(\lambda) \,, \\
        \frac{x - a(\lambda)}{b(\lambda) - a(\lambda)}, \,\text{if } a(\lambda) \leq x < b(\lambda) \,, \\
        1, \,\text{if } b(\lambda) \leq x \,,
    \end{cases}
\end{split}
\end{equation}
which has derivative $(D\lam{\mu} \ast \sigma)(x) = \lam{\mu}(x)$.
Since $\alpha(\lambda), \beta(\lambda) \to 0$ as $\lambda \to 0$, $\lam{\mu} \to \delta_{0}$ (the Dirac's delta centred at zero) in the distributional sense, and $\lam{\sigma} \to \sigma$ in the pointwise sense as requested by our definition of regularised function.

How does this discussion connect with \gls{STE}?
If we set $\alpha(\lambda) \equiv 1 / 2$ and $\beta(\lambda) \equiv 1 / 2 \sqrt{3}$ in \eqref{eq:ste_uniform}, $\lam{\sigma}$ is the clipped ReLU.
Similarly, if we set $\alpha(\lambda) \equiv 0$ and $\beta(\lambda) \equiv 1 / 2 \sqrt{3}$, $\lam{\sigma}$ is the hard sigmoid.
The same principle can be adapted to derive the piecewise polynomial regularisation of $H^{+}_{0}$ proposed in \cite{Deng2018, Liu2018} (corresponding to a triangular noise distribution), as well as the error function (in case of normal noise) and the logistic function (in case of logistic noise).
The observation that all these functions can be seen as regularisations of the Heaviside is not new \cite{Wu2018, Severa2019}; however, Proposition~\ref{th:regularisation} generalises to broader classes of regularisations.

\subsection{Dynamic \gls{STE} and compositional convergence}

In this sub-section, we consider the problem of regularising \glspl{QNN} with \gls{STE} variants that can evolve through time.
First, we will set the formalism to describe arbitrary neural networks; after defining compositions of quantised layer maps, we will define compositions of regularised layer maps; finally, we will define the concept of compositional convergence and briefly discuss its implications

Let $L \geq 2$ be an integer \textbf{number of layers}.
Given an integer \textbf{input size} $n_{0} \geq 1$, let $X_{0} \subset \R^{n_{0}}$ be the \textbf{input space}.
For each $\l = 1, \dots, L$, define an integer \textbf{layer size} $n_{\l} \geq 1$, a \textbf{feature space} $X_{\l} \subseteq \R^{n_{\l}}$, a space of \textbf{weight matrices} $W_{\l} \subseteq \R^{n_{\l} \times n_{\l-1}}$, a space of \textbf{bias vectors} $B_{\l} \subseteq \R^{n_{\l}}$, and the \textbf{parameter space} $M_{\l} \coloneqq W_{\l} \times B_{\l}$.

For each $\l = 1, \dots, L$, given a fixed $\m_{\l} = (\W_{\l}, \b_{\l}) \in M_{\l}$, define the $\l$-th \textbf{layer map}
\begin{equation}\label{eq:layer_map}
    \ph_{\m_{\l}} \coloneqq \bm{\sigma}_{\l} \circ S_{\m_{\l}}
\end{equation}
as the composition of the affine map
\begin{equation}\label{eq:layer_map_affine}
\begin{split}
    S_{\m_{\l}} \,:\,
    X_{\l-1}  &\to \R^{n_{\l}} \\
    \x_{\l-1} &\mapsto \W_{\l} \x_{\l-1} + \b_{\l} \eqqcolon \s_{\l} \,,
\end{split}
\end{equation}
and the element-wise map
\begin{equation}\label{eq:layer_map_activation}
\begin{split}
    \bm{\sigma}_{\l} \,:\,
    \R^{n_{\l}} &\to X_{\l} \\
    \s_{\l}     &\mapsto \bm{\sigma}_{\l}(\s_{\l}) = (\sigma_{\l, 1}(s_{\l, 1}), \dots, \sigma_{\l, n_{\l}}(s_{\l, n_{\l}}))' \eqqcolon \x_{\l} \,,
\end{split}
\end{equation}
where the functions $\sigma_{\l, i_{\l}} : \R \to \R, i_{\l} = 1, \dots, n_{\l}$, are \textbf{activation functions}.
Activation functions are assumed to be non-constant and non-decreasing.
For layer maps $\l = 1, \dots, L-1$, there must be at least one activation function $\sigma_{\l, i_{\l}}$ which is non-linear (e.g., a bounded function).
Both in theory and applications it is usually assumed that $\bm{\sigma}_{L}$ is the identity function.

Given $\bl \in \{ 1, \dots, L \}$, we define $\hm_{\bl} \coloneqq (\m_{1}, \dots, \m_{\bl})$ to be the collective parameter taken from $\hM_{\bl} \coloneqq M_{1} \times \dots \times M_{\bl}$.
We define a \textbf{network map} recursively as follows:
\begin{equation}\label{eq:network_map}
\begin{split}
    \Phi_{\hm_{1}}  &\coloneqq \ph_{\m_{1}} \,, \\
    \Phi_{\hm_{\l}} &\coloneqq \ph_{\m_{\l}} \circ \Phi_{\hm_{\l-1}} \,,\, \l = 2, \dots, L \,.
\end{split}
\end{equation}

Consider a network map $\Phi_{\hm_{L}}$ such that all its activation functions $\sigma_{\l, i_{\l}}, i_{\l} = 1, \dots, n_{\l}, \l = 1, \dots, L-1$ are the Heaviside \eqref{eq:heaviside}.
For each $\l = 1, \dots, L-1$, let $\sigma_{\lambda_{\l}} : \R \to \R$ be a parametric regularisation of \eqref{eq:heaviside} with regularisation parameter $\lambda_{\l} = \lambda_{\l}(\lambda) > 0$, and $\bm{\sigma}_{\lambda_{\l}} : \R^{n_{\l}} \to \R^{n_{\l}}$ be the component-wise application of $\sigma_{\lambda_{\l}}$.
Analogously to \eqref{eq:layer_map}, we define the $\l$-th \textbf{regularised layer map} as $\ph_{\lambda_{\l}, \m_{\l}} \coloneqq \bm{\sigma}_{\lambda_{\l}} \circ S_{\m_{\l}}$.
Analogously to \eqref{eq:network_map}, we define the \textbf{regularised network map} as
\begin{equation}\label{eq:regularised_network_map}
\begin{split}
    \Phi_{\hat{\lambda}_{1}, \hm_{1}} &\coloneqq \ph_{\lambda_{1}, \m_{1}} \,, \\
    \Phi_{\hat{\lambda}_{\l}, \hm_{\l}} &\coloneqq \ph_{\lambda_{\l}, \m_{\l}} \circ \Phi_{\hat{\lambda}_{\l-1}, \hm_{\l-1}} \,,\, \l = 2, \dots, L \,.
\end{split}
\end{equation}

Given $\x_{0} \in X_{0}$, we define
\begin{equation}
    \x_{\l} \coloneqq \Phi_{\hm_{\l}}(\x_{0}) \,,\, \l = 1, \dots, L \,,
\end{equation}
to be the \textbf{quantised features} of $\x_{0}$ and
\begin{equation}
    \x_{\hat{\lambda}_{\l}, \l} \coloneqq \Phi_{\hat{\lambda}_{\l}, \hm_{\l}}(\x_{0}) \,,\, \l = 1, \dots, L \,.
\end{equation}
to be the \textbf{regularised features} of $\x_{0}$.
We are interested in understanding under which conditions on the evolution of $\sigma_{\lambda_{\l}}$ the \textbf{regularised feature} $\x_{\hat{\lambda}_{\l}, \l}$ converges to the \textbf{quantised feature} $\x_{\l}$.
We recall that a \textbf{convergence rate} is a continuous and non-decreasing function $r : \R^{+}_{0} \to \R^{+}_{0}$ satisfying $\lim_{\lambda \to 0} r(\lambda) = 0$.

\begin{theorem}
\label{th:compositional_convergence}
Consider a network map \eqref{eq:network_map} parametrised by $\hm^{L} = (\m^{1}, \m^{2}, \dots, \m^{L}) \in \hM^{L}$ and using the Heaviside $H^{+}_{0}$ as its activation function.

Consider a regularised network map \eqref{eq:regularised_network_map} such that for $\l = 1, \dots, L-1$ the regularisations $\sigma_{\lambda_{\l}}$ of $H^{+}_{0}$ and the regularisation parameters $\lambda_{\l}$ satisfy the following conditions:
\begin{align}
    &\lambda_{\l} \xrightarrow[\lambda \to 0]{} 0 \,; \label{eq:regparam_hp_i} \\
    &\sigma_{\lambda_{\l}}(s) \xrightarrow[\lambda_{\l} \to 0]{} H^{+}_{0}(s) \,,\, \,\forall\, s \in \R \,; \label{eq:regparam_hp_ii} \\
    &\text{$\sigma_{\lambda_{\l}}$ is strictly increasing} \,; \label{eq:regparam_hp_iii} \\
    &0 \leq \sigma_{\lambda_{\l}}(s) \leq 1 \,,\, \,\forall\, s \in \R \,. \label{eq:regparam_hp_iv}
\end{align}

Additionally, we assume that convergence rates $r_{\l}(\lambda), \l = 1, \dots, L$ are given, such that for every $\eps > 0$
\begin{eqnarray}
    \label{eq:convergence_hp_i} \sigma_{\lambda_{\l}}^{-1}(\eps r_{\l}(\lambda)) \xrightarrow[\lambda \to 0]{} 0 \,,& \\
    \label{eq:convergence_hp_ii} \sigma_{\lambda_{\l}}^{-1}(1 - \eps r_{\l}(\lambda)) \xrightarrow[\lambda \to 0]{} 0 \,,& \\
    \label{eq:convergence_hp_iii} \frac{1 - \sigma_{\lambda_{\l}}(0)}{r_{\l}(\lambda)} \xrightarrow[\lambda \to 0]{} 0 \,,&
\end{eqnarray}    
and, for $\l = 2, \dots, L$,
\begin{equation}\label{eq:convergence_hp_iv}
    \frac{r_{\l-1}(\lambda)}{\sigma_{\lambda_{\l}}^{-1}(1 - \eps r_{\l}(\lambda))} \xrightarrow[\lambda \to 0]{} 0 \,.
\end{equation}

Then, for any given $\x^{0} \in X^{0}$, we have
\begin{equation}\label{eq:convergence_thesis}
    \frac{\| \x_{\hat{\lambda}_{\l}, \l} - \x_{\l} \|}{r_{\l}(\lambda)} \xrightarrow[\lambda \to 0]{} 0 \,, \qquad\forall\, \l = 1, \dots, L \,.
\end{equation}
\end{theorem}
If \eqref{eq:regparam_hp_i}-\eqref{eq:convergence_hp_iv} hold, we say that the regularisations $\sigma_{\lambda_{\l}}, \l = 1, \dots, L$ satisfy the \textbf{compositional convergence} hypothesis.

In other words, the theorem states that parametric regularisations should converge quantitatively faster for the layer maps closer to the input.
Although this result describes how the regularised network processes information in the forward direction and can not explain the coherence between the regularised gradient (what is referred to as "coarse gradient" in \cite{Yin2019}) and the population loss gradient, it provides a partial justification for the empirical findings of \cite{Severa2019}.
Indeed, the Whetstone method uses "annealing schedules" where the regularisations are stabilised in earlier layer maps before proceeding to later ones.

\subsection{A connection between \gls{STE} targets and \gls{STE} regularisations}

Consider an input $x$ to a quantiser \eqref{eq:quantiser}.
When $x$ is added noise following a distribution $\lam{\mu}$, the probability of sampling the $k$-th quantisation level is $p_{k} = \mu_{\lambda, x}(\sigma^{-1}(\{ q_{k} \})) = \mu_{\lambda, x}(I_{k})$, where $\mu_{\lambda, x}(A) \coloneqq \lam{\mu}(A - x), \,\forall\, A \in \mathcal{A}$ ($\mathcal{A}$ is the Borel $\sigma$-algebra on $\R$).
When $\lam{\mu}$ is a zero-mean, uni-modal noise distribution, selecting the mode quantisation level $q_{k}' = \argmax_{q_{k} \in Q} \mu_{\lambda, x}(I_{k})$ returns the same value as applying the original $\sigma$ directly to $x$.
This observation is a more formal definition of what was referred to as "deterministic sampling" in \cite{Hubara2016}; we will make use of it in the next section.


\section{The \gls{ANA} algorithm}
\label{sec:ana}

Consider the main propositions of Section~\ref{sec:formulation}:
\begin{itemize}
    \item different \gls{STE} regularisations can be modelled as the expected value of the corresponding \gls{STE} target when it processes stochastic inputs;
    \item supposing that the noise distributions governing the parametric \gls{STE} regularisations in a given feedforward network can evolve dynamically, we can ensure that the composition of the regularisations pointwise converges to the composition of the \gls{STE} targets by enforcing the compositional convergence hypothesis;
    \item the \gls{STE} target and \gls{STE} regularisation can be seen as the same stochastic function operating according to two different strategies: mode in the forward pass, and expectation in the backward pass.
\end{itemize}

Based on these observations, we propose \gls{ANA}, whose pseudo-code is listed in Algorithm~\ref{algo:ana}.
It takes in input an \gls{STE}-regularised network map \eqref{eq:regularised_network_map} initialised at $\hm_{L}^{(0)}$; a schedule $\mathcal{S}$ mapping tuples $(\l, t)$ to the regularisation parameter $\lambda_{\l}$ for the $\l$-th layer at the $t$-th iteration; a training data set $\mathcal{D} = ((x^{(1)}, y^{(1)}), \dots, (x^{(N)}, y^{(N)}))$, $N \geq 1$ being an integer number of data points; a loss function $\mathcal{L} : X_{L} \times X_{L} \to \R^{+}_{0}$; a learning rate $\eta \in \R^{+}$; an integer $T_{e} \geq 1$ number of training epochs.
The routine $\mathtt{set\_noise}$ performs look-ups from the schedule $\mathcal{S}$, whereas the routine $\mathtt{optim}$ computes the parameter updates.
The total number of training iterations is $T \coloneqq T_{e} N$.
Of course, one can perform mini-batch \gls{SGD} (mini-batch size greater than one) instead of vanilla \gls{SGD}.

\begin{algorithm}\captionsetup{labelsep=newline}
\caption{%
\begin{tabular}{ll}
    \textbf{Input: } &$\Phi_{\hat{\lambda}_{L}, \hm_{L}^{(0)}}, \mathcal{S}, \mathcal{D}, \mathcal{L}, \eta, T_{e}$ \\
    \textbf{Output: } &$\Phi_{\hat{\lambda}_{L}, \hm_{L}^{(T)}}$ \\
    \textbf{Uses: } &$\mathtt{set\_noise}, \mathtt{optim}$
\end{tabular}
}
\begin{algorithmic}[1]
    \State $t \gets 0$
    \For{$t_{e} \gets 1:T_{e}$}
        \For{$(x, y') \,\mathtt{in}\, \mathcal{D}$}
            \State $t \gets t + 1$
            \For{$\l \gets 1:L$} \label{algo:ana:set_noise_s}
                \State $\lambda_{\l} \gets \mathtt{set\_noise}(\mathcal{S}, \l, t)$
            \EndFor \label{algo:ana:set_noise_e}
            \State $y \gets \Phi_{\hat{\lambda}_{L}, \hm_{L}^{(t-1)}}(x)$ \Comment{inference}\label{algo:ana:inference}
            \State $\g^{(t)} \gets \nabla_{\hm_{L}} \mathcal{L}(y, y')$ \Comment{backpropagation}\label{algo:ana:backpropagation}
            \State $\hm_{L}^{(t)} \gets \mathtt{optim}(\hm_{L}^{(t-1)}, \g^{(t)}, \eta, t)$ \Comment{update}
        \EndFor
    \EndFor
    \State \Return $\Phi_{\hat{\lambda}_{L}, \hm_{L}^{(t)}}$
\end{algorithmic}
\label{algo:ana}
\end{algorithm}

The first hyper-parameter of \gls{ANA} is the collection $(\mu_{\lambda_{1}}, \dots, \mu_{\lambda_{L}})$ of parametric probability measures used to regularise the layer maps $\ph_{\lambda_{\l}, \m_{\l}}, \l = 1, \dots, L$.

The second hyper-parameter is the evolution of the regularisations as the training algorithm progresses.
This hyper-parameter is encoded in the schedule $\mathcal{S}$ and controlled by Lines~\ref{algo:ana:set_noise_s}-\ref{algo:ana:set_noise_e}.
Note that static schedules are also allowed as special cases.

We also encoded a third hyper-parameter, which is not explicitly reported in Algorithm~\ref{algo:ana}: the forward computation strategy.
Given a layer map using a quantiser \eqref{eq:quantiser}, we allow the forward pass to use either the expected value of the regularised quantiser, $\E_{\mu_{\lambda_{\l}}}[\sigma_{\nu}]$, the mode $q_{k}' = \argmax_{q_{k} \in Q} (p_{k} \coloneqq \mu_{\lambda_{\l}, x}(I_{k}))$, or a random sampling $q_{k}' \sim ((p_{0}, \dots, p_{K-1}), 1)$.
We name these strategies expectation, mode (or \textbf{deterministic sampling}) and random (or \textbf{random sampling}).

\section{Experimental results}
\label{sec:experiments}

\subsection{Methodology and experiment design}

CIFAR-10 is a popular small data set for image classification \cite{Krizhevsky2014}.
It contains $60k$ RGB-encoded images partitioned into ten semantic classes.
It consists of a training partition containing $5k$ images per class and a validation partition containing $1k$ images per class.

As a reference, we used a simple fully-feedforward network using five convolutional layers and three linear layers, inspired by the VGG family of topologies \cite{Simonyan2015}; therefore, $L = 8$.
We quantised all the weights and features to be ternary, apart from the last layer, which we kept in floating-point format coherently with common literature practice \cite{Choi2019, Esser2020}.

In each experimental unit, we trained the network for $500$ epochs using mini-batches of $256$ images, the cross-entropy loss function, and the ADAM optimiser with an initial learning rate of $10^{-3}$, decreased to $10^{-4}$ after $400$ epochs.

Our experimental design consists of six \glspl{DoF}: noise type, static mean, static variance, decay interval, decay power law, forward computation strategy.
We detail their purpose in the following paragraphs.

The \textbf{noise type} is the parametric family of distributions used to instantiate $\mu_{\lambda_{\l}}, \l = 1, \dots, L - 1$.
We considered four noise types: \textbf{uniform}, \textbf{triangular}, \textbf{normal} and \textbf{logistic}.
We parametrised the distribution densities $\mu_{\lambda_{\l}}$ in their means $\alpha(\lambda_{\l})$ and standard deviations $\beta(\lambda_{\l})$:
\begin{align*}
    \mu_{\lambda_{\l}}(x) &= \frac{\chi_{[\alpha(\lambda_{\l}) - \sqrt{3} \beta(\lambda_{\l}), \alpha(\lambda_{\l}) + \sqrt{3} \beta(\lambda_{\l})]}(x)}{2 \sqrt{3} \beta(\lambda_{\l})} \,; \\
    \mu_{\lambda_{\l}}(x) &=
    \begin{cases}
        0, \text{if } x \notin [\alpha(\lambda_{\l}) - \sqrt{6} \beta(\lambda_{\l}), \alpha(\lambda_{\l}) + \sqrt{6} \beta(\lambda_{\l})) \,, \\
        \frac{x - \alpha(\lambda_{\l}) + \sqrt{6} \beta(\lambda_{\l})}{6 \beta^{2}(\lambda_{\l})}, \text{if } x \in [\alpha(\lambda_{\l}) - \sqrt{6} \beta(\lambda_{\l}), \alpha(\lambda_{\l})) \,, \\
        \frac{\alpha(\lambda_{\l}) + \sqrt{6} \beta(\lambda_{\l}) - x}{6 \beta^{2}(\lambda_{\l})}, \text{if } x \in [\alpha(\lambda_{\l}), \alpha(\lambda_{\l}) + \sqrt{6} \beta(\lambda_{\l})) \,;
    \end{cases} \\
    \mu_{\lambda_{\l}}(x) &= \frac{e^{-\frac{(x - \alpha(\lambda_{\l}))^{2}}{2 \beta^{2}(\lambda_{\l})}}}{\sqrt{2 \pi} \beta(\lambda_{\l})} \,; \\
    \mu_{\lambda_{\l}}(x) &= \frac{e^{-\frac{x - \alpha(\lambda_{\l})}{\beta(\lambda_{\l})}}}{\beta(\lambda_{\l}) \left( 1 + e^{-\frac{x - \alpha(\lambda_{\l})}{\beta(\lambda_{\l})}} \right)^{2}} \,.
\end{align*}
Note that all these distributions are uni-modal and have densities that are symmetric with respect to the mean.

Also, note that the uniform and triangular distributions' densities have compact support, whereas the normal and logistic have non-zero densities over all $\R$.
To compare the regularisations obtained using compactly-supported and non-compactly-supported densities, we considered a non-compactly-supported measure $\mu_{1}$ as equivalent to a compactly-supported measure $\mu_{2}$ if they had the same mean and exactly $95\%$ of the total probability mass of $\mu_{1}$ fell in the support of $\mu_{2}$.

The \textbf{noise schedule} defines how the measures $\mu_{\lambda_{\l}}, \l = 1, \dots, L - 1$ evolve through time.
This evolution is controlled through the shape parameters $\alpha(\lambda_{\l}), \beta(\lambda_{\l})$.
In particular, denoting by $\alpha(\lambda_{\l}^{(t)})$ and $\beta(\lambda_{\l}^{(t)})$ the parameters of the distribution $\mu_{\lambda_{\l}}$ at the $t$-th iteration, their values are determined by the following functions:
\begin{align*}
    \alpha(\lambda_{\l}^{(t)}) = c_{\alpha, \l} f_{\alpha, \l}(\lambda_{\l}^{(t)}) \,, \\
    \beta(\lambda_{\l}^{(t)}) = c_{\beta, \l} f_{\beta, \l}(\lambda_{\l}^{(t)}) \,.
\end{align*}
Here, $c_{\alpha, \l}, c_{\beta, \l}$ are non-negative real constants.

The \textbf{static mean} \gls{DoF} is a Boolean variable.
When it is set to true, then $f_{\alpha, \l}(\lambda_{\l}^{(t)}) \equiv 1$ and $c_{\alpha, \l} = 0$.
When it is set to false, then $c_{\alpha, \l} > 0$ and
\begin{equation}\label{eq:mean_evolution}
\begin{split}
    f_{\alpha, \l}(\lambda_{\l}^{(t)})
    &= (\lambda_{\l}^{(t)})^{d_{\alpha, \l}} \\
    &= \max\left\{ 0, \min\left\{ \left( \frac{t_{\l, end} - t}{t_{\l, end} - t_{\l, start}} \right), 1 \right\} \right\}^{d_{\alpha, \l}}
\end{split}
\end{equation}
for some $0 \leq t_{\l, start} < t_{\l, end} \leq T$ and an integer $d_{\alpha, \l} \geq 1$.

Similarly, the \textbf{static variance} \gls{DoF} is a Boolean variable such that $f_{\beta, \l}(\lambda_{\l}^{(t)}) \equiv 1$ when it is set to true, and
\begin{equation}\label{eq:std_evolution}
\begin{split}
    f_{\beta, \l}(\lambda_{\l}^{(t)})
    &= (\lambda_{\l}^{(t)})^{d_{\beta, \l}} \\
    &= \max\left\{ 0, \min\left\{ \left( \frac{t_{\l, end} - t}{t_{\l, end} - t_{\l, start}} \right), 1 \right\} \right\}^{d_{\beta, \l}}
\end{split}
\end{equation}
for some integer $d_{\beta, \l} \geq 1$ when it is set to false.
We always consider $c_{\beta, \l} > 0$, since setting it to zero amounts to no regularisation and, therefore, zero gradients throughout the training process, resulting in the target network not being capable of learning.

When the static variance \gls{DoF} is set to false, we assume that each distribution $\mu_{\lambda_{\l}}$ is annealed from some initial distribution $\mu_{\lambda_{\l}^{(0)}}$ to the final distribution $\mu_{\lambda_{\l}^{(T)}} = \delta_{0}$, which is a Dirac's delta centred at zero.
For each distribution, we define two instants $0 \leq t_{\l, start} < t_{\l, end} \leq T$ such that $\mu_{\lambda_{\l}^{(t)}} = \mu_{\lambda_{\l}^{(0)}}, 0 \leq t \leq t_{\l, start}$ and $\mu_{\lambda_{\l}^{(t)}} = \mu_{\lambda_{\l}^{(T)}}, t_{\l, end} \leq t \leq T$.
We call the sequence $\{ t_{\l, start}, \dots, t_{\l, end} \}$ the \textbf{annealing range} of $\mu_{\lambda_{\l}}$.

The \textbf{decay interval} \gls{DoF} is a categorical variable defining how the annealing ranges of the various measures $\mu_{\lambda_{\l}}$ are mutually related:
\begin{itemize}
    \item \textbf{same start}: $t_{\l, start}$ is the same for all layers, but $t_{\l-1, end} < t_{\l, end}$ for $\l = 2, \dots, L$;
    \item \textbf{same end}: $t_{\l, start} < t_{\l-1, start}$ for $\l = 2, \dots, L$, but $t_{\l, end}$ is the same for all layers;
    \item \textbf{partition}: $t_{\l-1, end} = t_{\l, start}$ for $\l = 2, \dots, L$;
    \item \textbf{overlapped}: $t_{\l_{1}, start} = t_{\l_{2}, start}$ and $t_{\l_{1}, end} = t_{\l_{2}, end}$ for all $\l_{1}, \l_{2} \in \{ 1, \dots, L \}$.
\end{itemize}

The \textbf{decay power law} \gls{DoF} is a binary categorical variable defining how the exponents $d_{\alpha, \l}, d_{\beta, \l}$ in \eqref{eq:mean_evolution} and \eqref{eq:std_evolution} relate to each other.
If it is set to \textbf{homogeneous}, then $d_{\alpha, \l}$ and $d_{\beta, \l}$ are the same for all the layers.
If it is set to \textbf{progressive}, then $d_{\alpha, \l}$ and $d_{\beta, \l}$ are inversely proportional to the depth of the layer: this option enforces a faster annealing of $\mu_{\lambda_{\l}}$ to $\delta_{0}$ over the corresponding decay range $\{ t_{\l, start}, \dots, t_{\l, end} \}$ in the layers that are closer to the input.
In other words: the decay interval controls the relative precedence of annealing ranges, whereas the decay power law determines the speed of decay.

The last \gls{DoF} of our experimental design is the \textbf{forward computation strategy}.
As discussed in Section~\ref{sec:ana}, it is a ternary categorical variable allowing the following options: \textbf{expectation}, \textbf{mode}, \textbf{random}.

The complete hyper-parameter search grid would consist of $384$ configurations.
However, we can avoid exploring a large subset of them by making some observations.

The combination of static mean and static variance determines the dynamics of the noise distributions; note that when both variables are set to true, then \gls{ANA} is equivalent to standard \gls{STE}.
The combination of decay interval and decay power law defines the annealing schedule; when the distributions are static, that is, when we are in the case of standard \gls{STE}, the value of this variable is not relevant, and we can fix it to an arbitrary value.
Out of $96$ experimental units that use static distributions, we need to explore only $12$.

We did not consider the case where the static variance variable is set in combination with the expectation forward computation strategy.
Indeed, the expected values of features and weights computed with respect to a non-collapsed noise distribution can greatly differ from the values computed with respect to a collapsed one; for this reason, the sudden removal of the noise at deployment time can disrupt the functional relationship learnt by the regularised network.
This observation allows us to cut additional $32$ out of $96$ experimental units that use dynamic noise, static variance, and the expectation forward computation strategy, as well as $4$ out of $12$ units that use static noise distributions.

To assign confidence intervals to our measurements, we evaluated each hyper-parameter configuration using five-fold cross-validation on the CIFAR-10 training partition.

\subsection{Results \& discussion}

\begin{figure}[t]
	\begin{subfigure}[b]{1.0\linewidth}
	    \centering\includegraphics[width=1.0\linewidth]{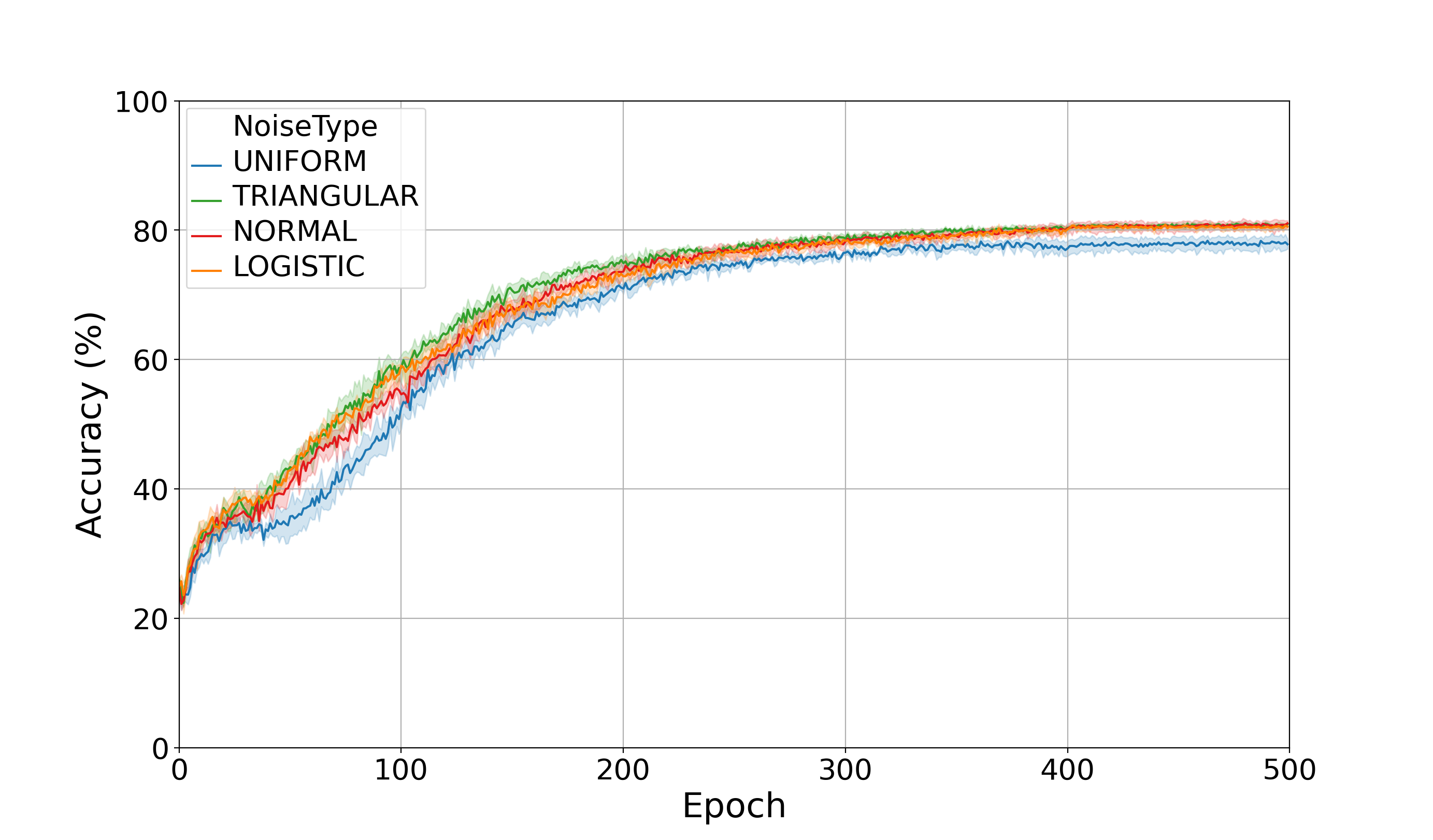}
	    \caption{}    \label{fig:cifar10_static_schedule_random}
	\end{subfigure}
	\begin{subfigure}[b]{1.0\linewidth}
	    \centering\includegraphics[width=1.0\linewidth]{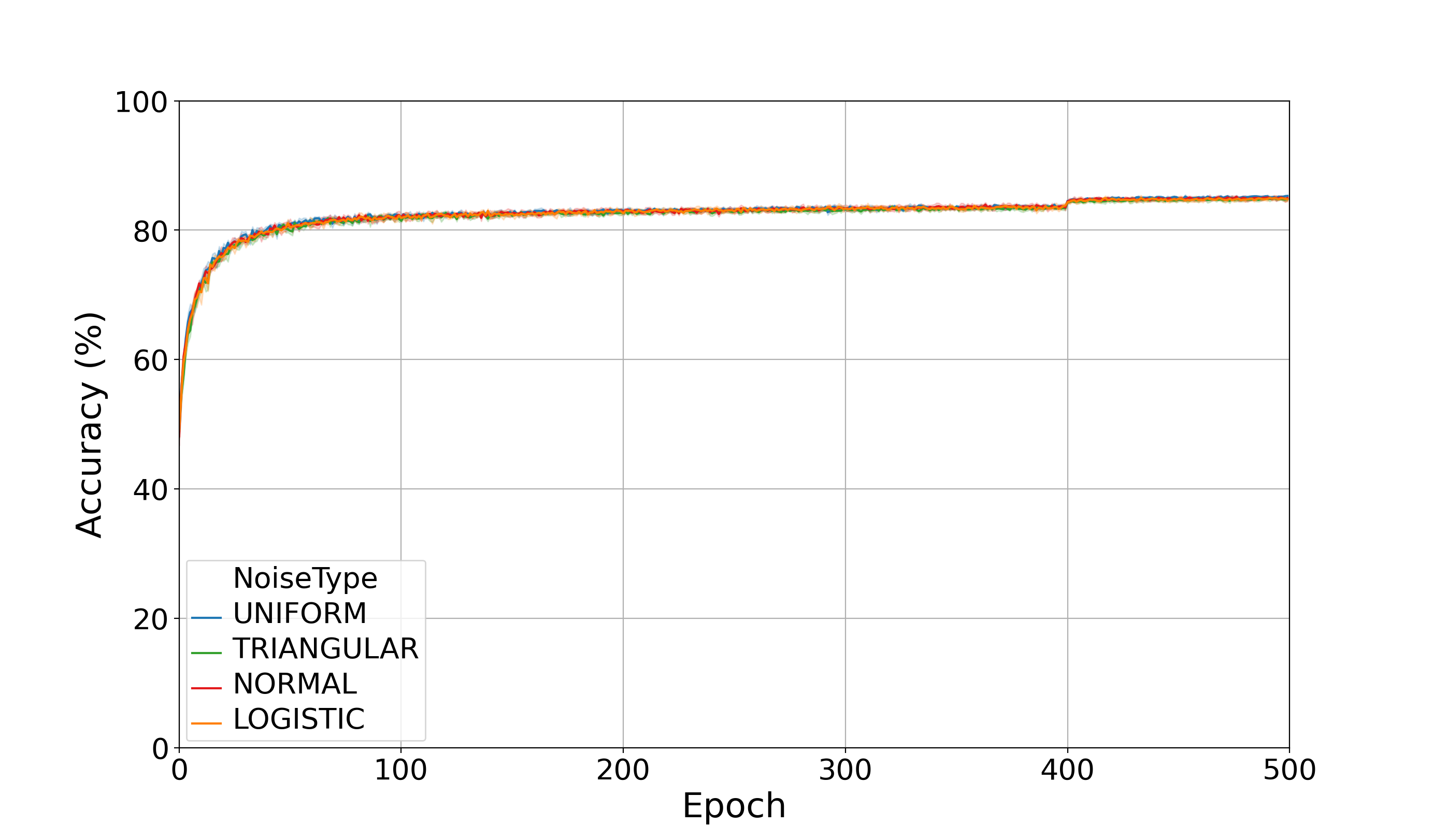}
	    \caption{}    \label{fig:cifar10_static_schedule_mode}
	\end{subfigure}
    \caption{Performance of \gls{ANA} using static noise schedules in combination with different forward computation strategies: random~\ref{fig:cifar10_static_schedule_random}, mode~\ref{fig:cifar10_static_schedule_mode}. Different noise types are reported using different colours: uniform (blue), triangular (green), normal (red), logistic (yellow).}
\end{figure}

\textbf{Static noise schedules} (i.e., schedules that do not anneal the distributions $\mu_{\lambda_{\l}}$ to $\delta_{0}$) represent the baseline for our comparisons.
Figures~\ref{fig:cifar10_static_schedule_random},~\ref{fig:cifar10_static_schedule_mode} show that there is a small advantage in using the mode forward computation strategy ($\sim 86\%$) as opposed to the random forward computation strategy ($\sim 82\%$).
In part, this advantage is due to the higher sensitivity of the former strategy to the learning rate lowering that takes place at epoch $400$.

In general, the noise type has negligible impact on task accuracy.
There is an exception when using the random forward computation strategy: using the uniform noise yields slightly worse results than other distributions.

\begin{figure}[t]
	\vspace{-16pt}
	\begin{subfigure}[b]{1.0\linewidth}
	    \centering\includegraphics[width=0.9\linewidth]{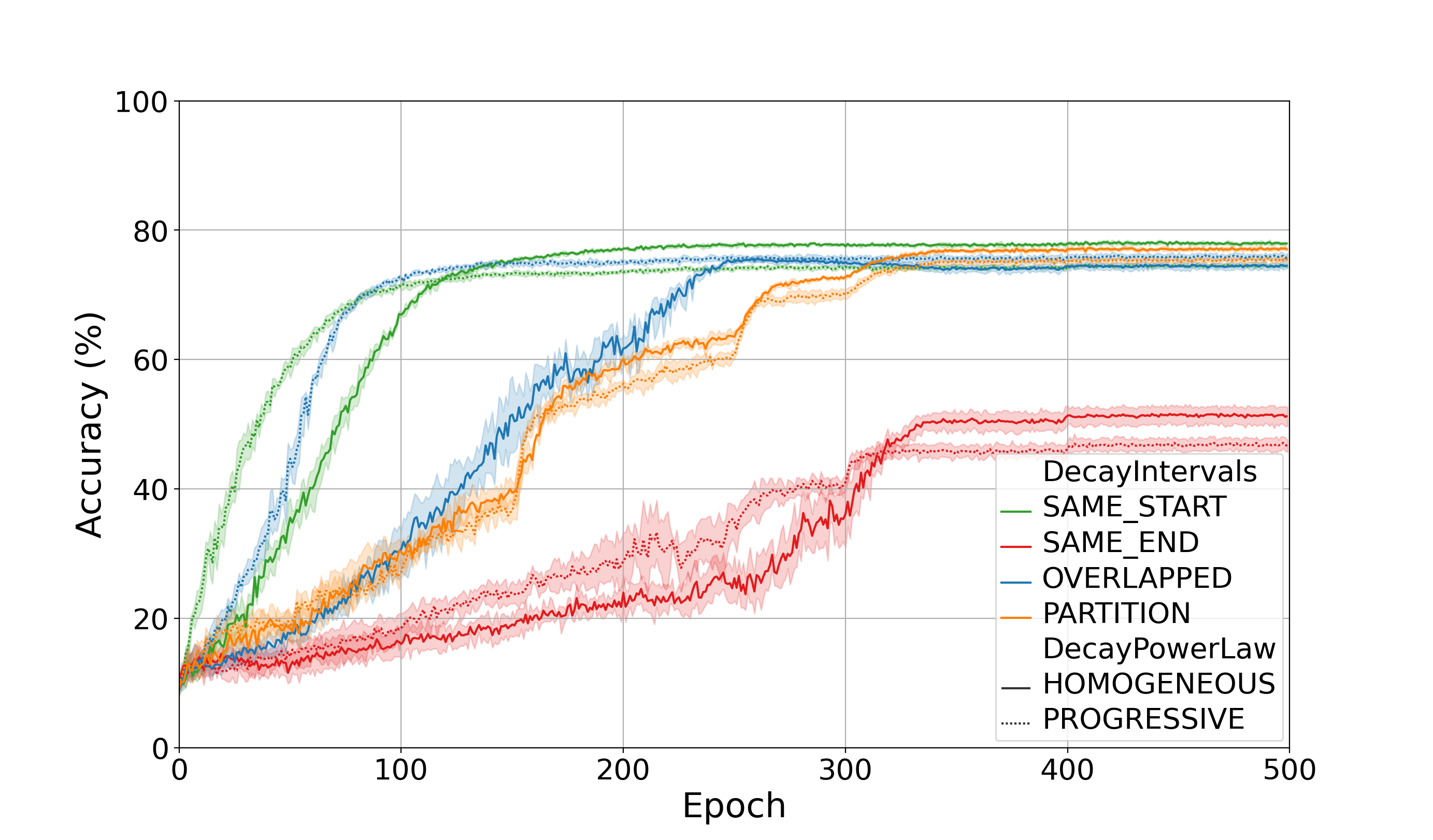}
    	\caption{}    \label{fig:cifar10_uniform_dynamic_schedule_expectation}
	\end{subfigure}
	\begin{subfigure}[b]{1.0\linewidth}
	    \centering\includegraphics[width=0.9\linewidth]{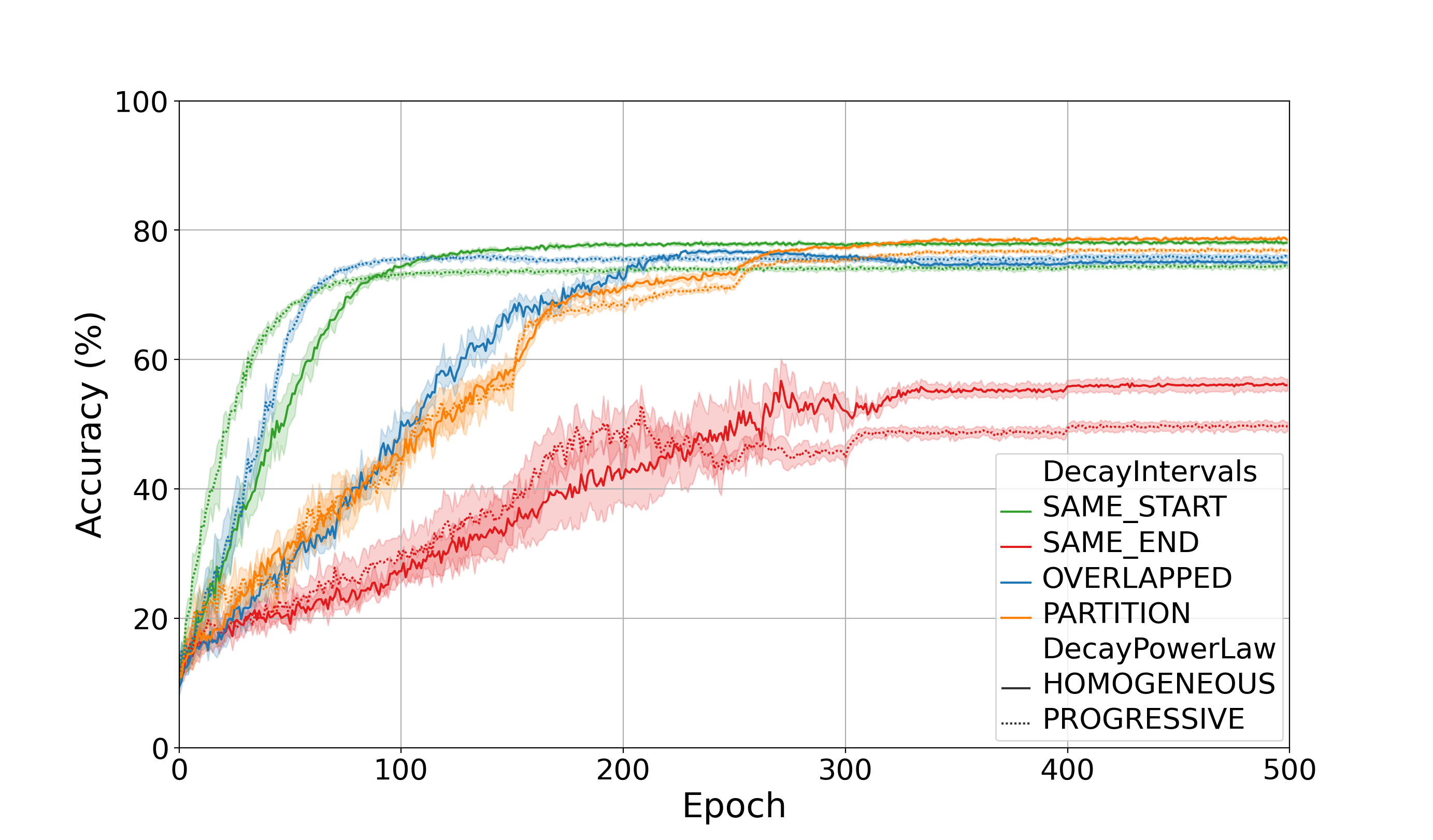}
	    \caption{}    \label{fig:cifar10_triangular_dynamic_schedule_expectation}
	\end{subfigure}
	\begin{subfigure}[b]{1.0\linewidth}
	    \centering\includegraphics[width=0.9\linewidth]{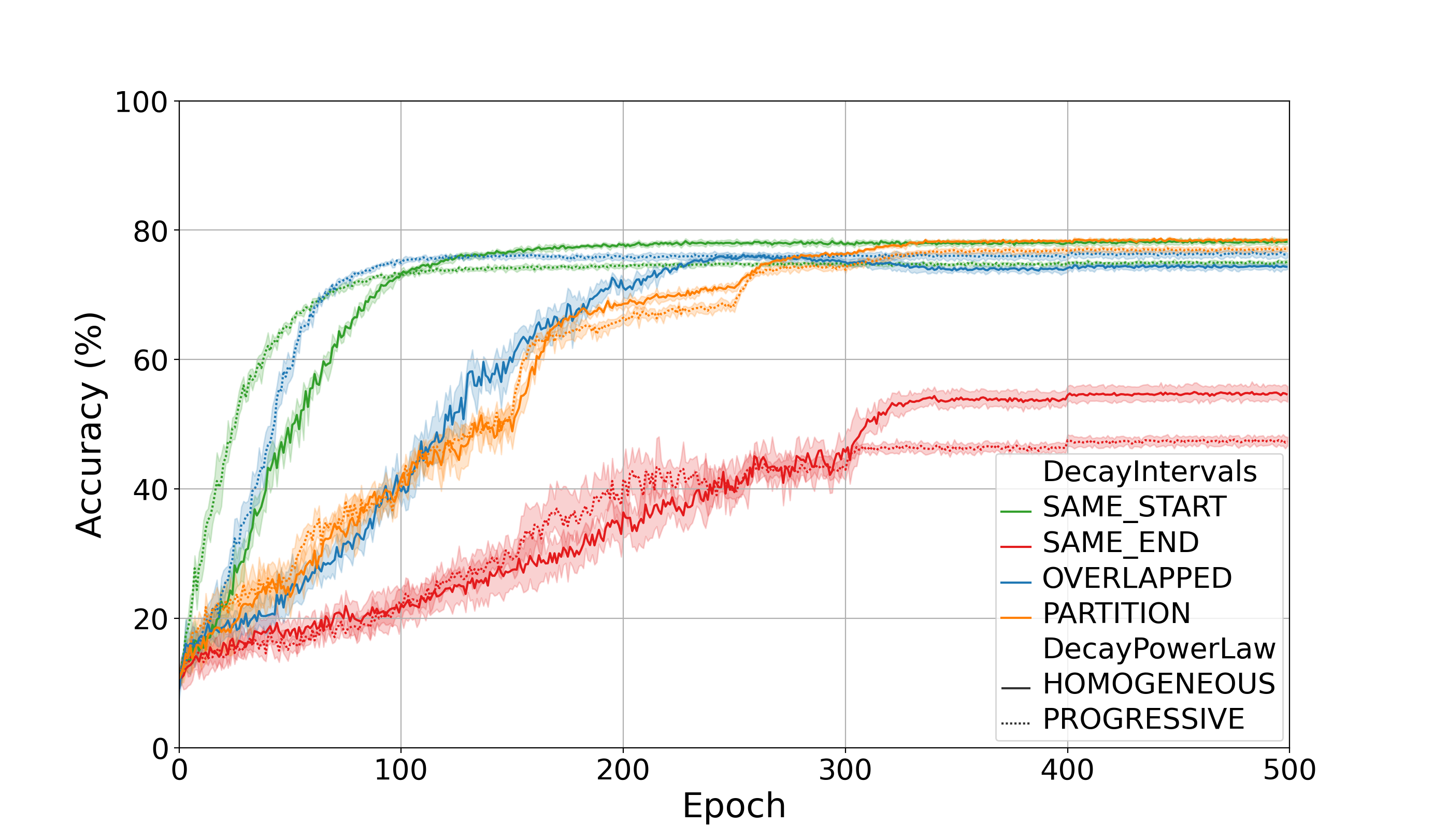}
	    \caption{}    \label{fig:cifar10_normal_dynamic_schedule_expectation}
	\end{subfigure}
	\begin{subfigure}[b]{1.0\linewidth}
	    \centering\includegraphics[width=0.9\linewidth]{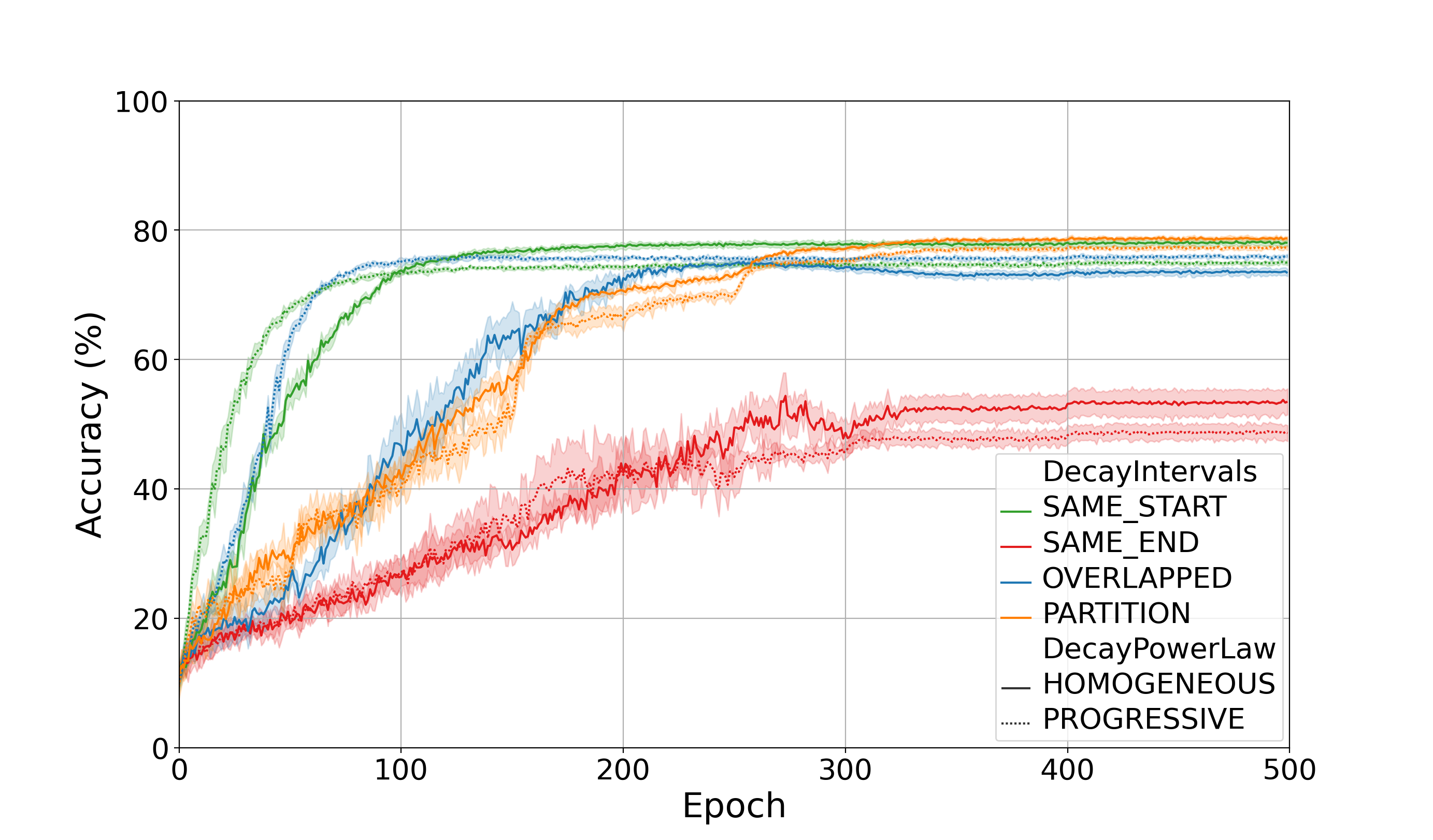}
    	\caption{}    \label{fig:cifar10_logistic_dynamic_schedule_expectation}
	\end{subfigure}
	\caption{Performance of \gls{ANA} using dynamic noise schedules under the expectation forward computation strategy and different noise types: uniform~\ref{fig:cifar10_uniform_dynamic_schedule_expectation}, triangular~\ref{fig:cifar10_triangular_dynamic_schedule_expectation}, normal~\ref{fig:cifar10_normal_dynamic_schedule_expectation}, logistic~\ref{fig:cifar10_logistic_dynamic_schedule_expectation}. Each plot reports multiple schedules: decay interval: same start (green), same end (red), partition (yellow), overlapped (blue); decay power law: homogeneous (continuous), progressive (dotted).}
\end{figure}

We start our analysis of \textbf{dynamic noise schedules} by considering the expectation forward computation strategy.
As can be seen in Figure~\ref{fig:cifar10_uniform_dynamic_schedule_expectation}, under a uniform noise distribution, the best convergence is achieved under the partition and same start decay interval strategies ($\sim 79\%$ accuracy), with a slight advantage given when using homogeneous decay power laws as opposed to the progressive ones.
However, the accuracy drop with respect to the static noise schedule baseline is not negligible ($-7\%$).
The same end decay interval strategy starts annealing the noise measure in the earlier layers only at a later moment during training, while the measures of the later layers are already converging, breaking the hypothesis of Theorem~\ref{th:compositional_convergence}: indeed, we see that the corresponding configuration of \gls{ANA} led to the worst performance over this batch of experimental units.

These observations were also confirmed for the triangular, normal, and logistic distributions, as shown in Figures~\ref{fig:cifar10_triangular_dynamic_schedule_expectation},~\ref{fig:cifar10_normal_dynamic_schedule_expectation},~\ref{fig:cifar10_logistic_dynamic_schedule_expectation}.
As for the static noise schedule case, the noise type does not have a major impact on accuracy.

So far, experimental evidence suggested that the noise type is not the most relevant variable.
Therefore, we fixed the noise type to uniform and analysed the impact of the forward computation strategy.

\begin{figure}[t]
	\begin{subfigure}[b]{1.0\linewidth}
	    \centering\includegraphics[width=1.0\linewidth]{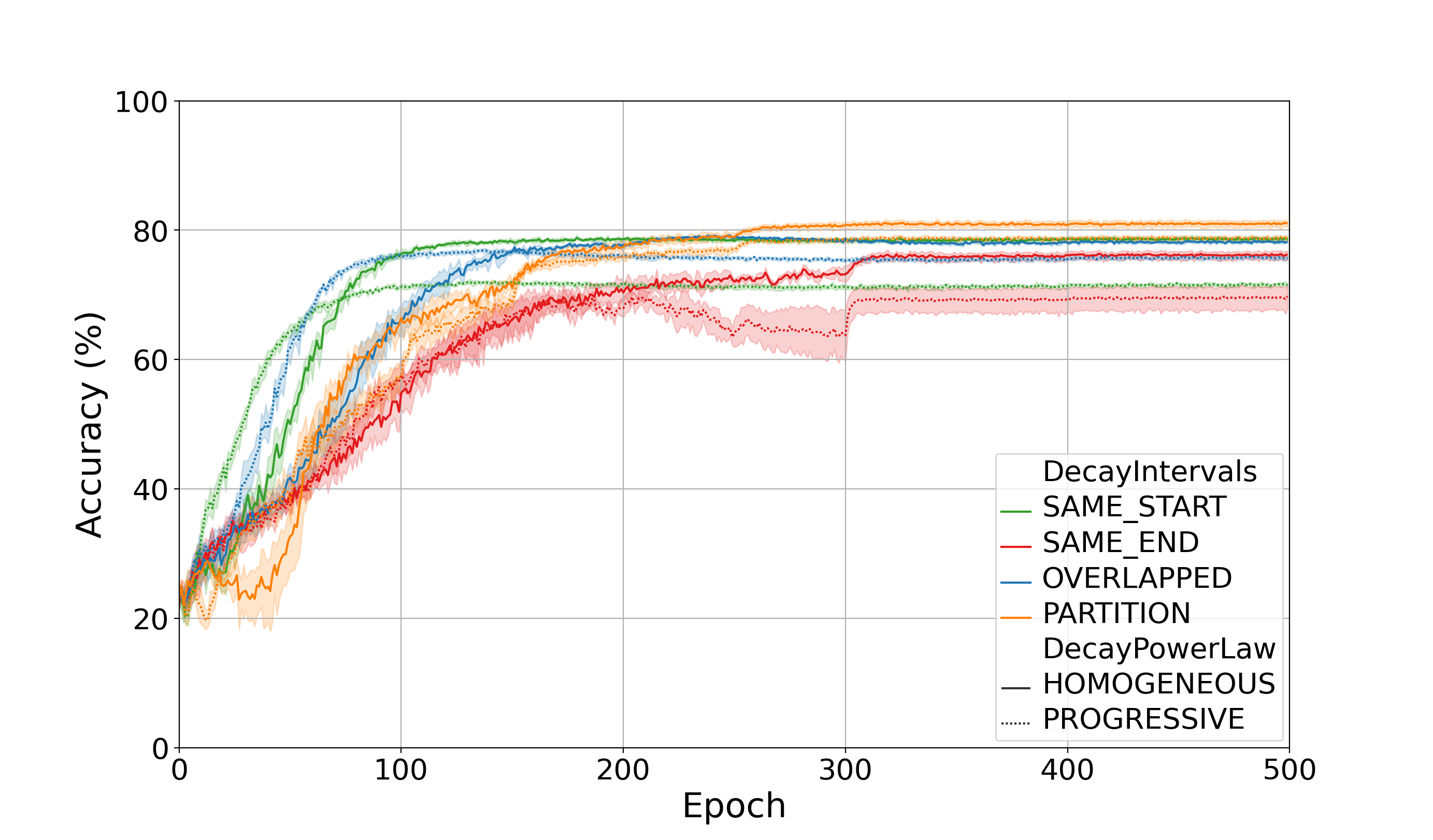}
    	\caption{}    \label{fig:cifar10_uniform_dynamic_schedule_random}
    \end{subfigure}
	\vspace{-2pt}
    \begin{subfigure}[b]{1.0\linewidth}
	    \centering\includegraphics[width=1.0\linewidth]{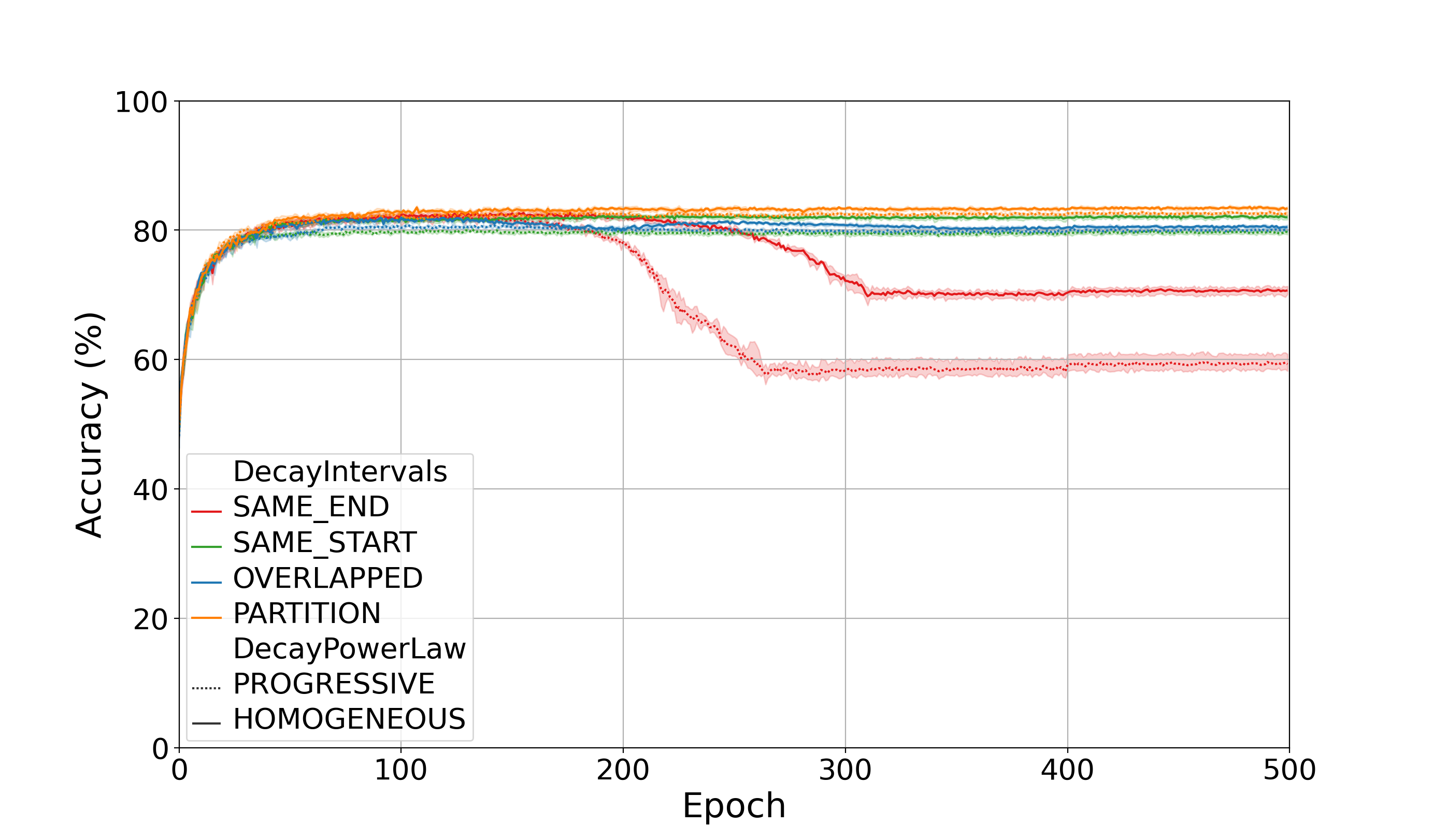}
    	\caption{}    \label{fig:cifar10_uniform_dynamic_schedule_mode}
    \end{subfigure}
    \caption{Performance of \gls{ANA} using dynamic noise schedules under the uniform noise type and different forward computation strategies: random~\ref{fig:cifar10_uniform_dynamic_schedule_random}, mode~\ref{fig:cifar10_uniform_dynamic_schedule_mode}. Each plot reports multiple schedules: decay interval: same start (green), same end (red), partition (yellow), overlapped (blue);  decay power law: homogeneous (continuous), progressive (dotted).}
\end{figure}

Figure~\ref{fig:cifar10_uniform_dynamic_schedule_random} shows the performance of \gls{ANA} under different schedules when random sampling is used in the inference pass.
We can see that random sampling combined with the partition decay range strategy can improve accuracy by $~4\%$.
Although random sampling seems to mitigate the degradation due to the inappropriate scheduling associated with the same end decay interval strategy, this strategy remains the worst in this batch of experimental units.

Using deterministic sampling in combination with the partition decay interval strategy seems to allow filling the gap with the networks trained using static schedules, as shown in Figure~\ref{fig:cifar10_uniform_dynamic_schedule_mode}.
We observe that the (small) advantage of static noise schedules is due to the learning rate lowering that takes place at epoch $400$; in the chosen dynamic noise schedule, the regularising noise distributions are annealed by epoch $350$, preventing gradients from tuning the parameters of all but the last layer.
However, the two noise schedules yield the same accuracy until the learning rate lowering.

\section{Conclusions}
\label{sec:conclusions}

Propagating gradients through discontinuities remains a critical problem in training quantised neural networks.
In this paper, we provided a unified framework to reason about STE and its variants.
We formally showed how stochastic regularisation can be used to derive entire families of STE variants.
Moreover, we analysed how dynamic \gls{STE} variants can be used to regularise discontinuous networks and showed how to properly synchronise them to guarantee convergence to the target function during the inference pass.
Our experiments on the CIFAR-10 benchmark highlighted that the major impact on accuracy is not due to the qualitative shape of the regularisers, but instead to the proper synchronisation of the STE variants used at different layers.
In particular, we observed a remarkable correspondence between the predictions of Theorem~\ref{th:compositional_convergence} and the empirical performance of different dynamic noise schedules.
Indeed, delaying the annealing of the regularising distributions in the earlier layers with respect to the later layers results in huge accuracy degradations.
Note that when the noise distributions are annealed to Dirac's deltas, local gradients evaluate to zero; therefore, gradient computation can be stopped whenever it reaches a quantiser with annealed noise since its output will not contribute to any update.
This observation implies that gradient propagation towards upstream nodes can be interrupted early during training when using the effective partition decay intervals strategy.
Consequently, ANA can be used to reduce the computational cost of the backward pass in QNN training, with possible benefits for on-chip training.

\paragraph{Acknowledgements}
We acknowledge the CINECA award under the ISCRA initiative, for the availability of high performance computing resources and support.  


\appendix

\clearpage
\section{Proofs}
\label{app:proofs}

\begin{lemma}\label{th:derivative}
Let $\sigma \,:\, \R \to \R$ and $\mu \,:\, \R \to \R$ be given functions.
The following facts hold:
\begin{itemize}
    \item[(i)] if $\sigma \in L^{\infty}(\R)$ and $\mu \in BV(\R)$ then $\mu \ast \sigma \in W^{1, \infty}(\R)$;
    \item[(ii)] if $\sigma \in L^{\infty}(\R)$ and $\mu \in W^{1, 1}(\R)$ then the weak derivative of $\mu \ast \sigma$ satisfies
    \begin{equation*}
        D(\mu \ast \sigma)(x) = (D\mu \ast \sigma)(x)
    \end{equation*}
    for almost all $x \in \R$;
    \item[(iii)] if $\sigma \in BV(\R)$ and $\mu \in W^{1, 1}(\R)$ then $\mu \ast \sigma \in C^{1}(\R)$, its derivative is uniformly continuous and one has
    \begin{equation*}
        \frac{d(\mu \ast \sigma)}{dx}(x) = (D\mu \ast \sigma)(x)
    \end{equation*}
    for all $x \in \R$.
\end{itemize}
\begin{proof}
We first show (i).
It is immediate to check that
\begin{equation}\label{eq:bound}
    \| \mu \ast \sigma \|_{\infty} \le \| \mu \|_{1} \, \| \sigma \|_{\infty} < +\infty \,.
\end{equation}
We choose $x_{1} < x_{2} \in \R$ and, thanks to Fubini's theorem and a change of variable, we obtain the following estimate:
\begin{equation*}
\begin{split}
    &| (\mu \ast \sigma)(x_{1}) - (\mu \ast \sigma)(x_{2}) | \\
    &\qquad= \left| \int_{y \in \R} \big( \mu(x_{1} - y) - \mu(x_{2} - y) \big) \sigma(y) \, dy \right| \\
    &\qquad\le \int_{y \in \R} |D\mu|\big( [x_{1} - y, x_{2} - y] \big) \, | \sigma(y) | \, dy \\
    &\qquad\le \| \sigma \|_{\infty} \int_{y \in \R} \int_{z \in \R} \chi_{[x_{1}, x_{2}]}(z + y) \, d|D\mu|(z) \, dy \\
    &\qquad= \| \sigma \|_{\infty} \int_{z \in \R} \int_{y \in \R} \chi_{[x_{1}, x_{2}]}(z + y) \, dy \, d|D\mu|(z) \\
    &\qquad\le \| \sigma \|_{\infty} \, |D\mu|(\R) \, |x_{1} - x_{2}| \,.
\end{split}
\end{equation*}
This shows that $\mu \ast \sigma$ is a Lipschitz function (with Lipschitz constant bounded above by $\| \sigma \|_{\infty} \, |D\mu|(\R)$).
Therefore the proof of (i) follows from the Sobolev characterisation of Lipschitz functions combined with \eqref{eq:bound}.
Let us prove (ii) by showing that $\mu \ast \sigma$ is weakly differentiable, thus providing a pointwise almost everywhere representation of its weak derivative.
Let $\phi \in C^{\infty}_{c}(\R)$ be a given test function.
By using Fubini's Theorem, the definition of weak derivative, and the change of variable in the integration, we obtain
\begin{equation*}
\begin{split}
    &\int_{\R} (\mu \ast \sigma)(x) \, \frac{d\phi}{dx}(x) \, dx \\
    &\qquad= \int_{x \in \R} \int_{y \in \R} \mu(x - y) \sigma(y) \, dy \, \frac{d\phi}{dx}(x) \, dx \\
    &\qquad= \int_{y \in \R} \int_{x \in \R} \mu(x - y) \frac{d\phi}{dx}(x) \, dx \, \sigma(y) \, dy \\
    &\qquad= -\int_{x \in \R} \int_{y \in \R} D\mu(x - y) \, \sigma(y) \, dy \, \phi(x) \, dx \\
    &\qquad= -\int_{x \in \R} (D\mu \ast \sigma)(x) \, \phi(x) \, dx \,.
\end{split}
\end{equation*}
This shows (ii).
We finally prove (iii).
By (i) and (ii) we already know that $\mu \ast \sigma \in W^{1, \infty}(\R)$ and its weak derivative satisfies $D(\mu \ast \sigma)(x) = (D\mu \ast \sigma)(x)$ for almost all $x \in \R$.
The conclusion is achieved as soon as we show that $(D\mu \ast \sigma)(x)$ is a continuous function.
We have for $x_{1} < x_{2} \in \R$
\begin{equation*}
\begin{split}
    &| (D\mu \ast \sigma)(x_{1}) - (D\mu \ast \sigma)(x_{2}) | \\
    &\qquad= \left| \int_{y \in \R} D\mu(x_{1} - y) \sigma(y) \, dy + \right. \\
    &\qquad\qquad \left. - \int_{y \in \R} D\mu(x_{2} - y) \sigma(y) \, dy \right| \\  
    &\qquad= \left| \int_{z \in \R} D\mu(z) \, \big( \sigma(x_{1} - z) - \sigma(x_{2} - z) \big) \, dz \right| \\
    &\qquad\le \int_{z \in \R} \int_{t \in \R} \chi_{[x_{1}, x_{2}]}(t + z) \, d|D\sigma|(t) \, |D\mu(z)| dz \\
    &\qquad= \int_{t \in \R} \int_{z \in \R} \chi_{[x_{1}, x_{2}]}(t + z) \, |D\mu(z)| dz \, d|D\sigma|(t) \\
    &\qquad= \int_{t \in \R} \int_{z \in \R} \chi_{[x_{1}, x_{2}]}(t + z) \, |D\mu(z)| dz \, d|D\sigma|(t) \,.
\end{split}
\end{equation*}
Denote by $\kappa$ the non-negative, finite Borel measure defined by $d\kappa = |D\mu(z)| dz$.
Since $\kappa$ is absolutely continuous with respect to the Lebesgue measure, for all $\epsilon > 0$ there exists $\delta > 0$ such that $|x_{1} - x_{2}| < \delta$ implies $\kappa([x_{1}, x_{2}]) < \epsilon$.
Therefore we get
\begin{equation*}
\begin{split}
    &| (D\mu \ast \sigma)(x_{1}) - (D\mu \ast \sigma)(x_{2}) | \\
    &\qquad\le \int_{t \in \R} \kappa([x_{1} - t, x_{2} - t]) \, d|D\sigma|(t) \\
    &\qquad\le \epsilon \, |D\sigma|(\R)
\end{split}
\end{equation*}
as soon as $|x_{1} - x_{2}| \le \delta$, which proves the uniform continuity of $D\mu \ast \sigma$ and concludes the proof.
\end{proof}
\end{lemma}

\subsection*{Proof of Proposition~\ref{th:regularisation}}
\begin{proof}
The first claim (i) follows from the definition of convolution.
The proofs of (ii) and (iii) follow from the application of Lemma~\ref{th:derivative}, noticing that, by definition, a quantiser \eqref{eq:quantiser} satisfies $\sigma \in L^{\infty}(\R)$ (in particular, $\| \sigma \|_{\infty} = \max_{q \in Q}\{ |q| \}$).
\end{proof}

\subsection*{Proof of Theorem~\ref{th:compositional_convergence}}
\begin{proof}
First, we note that
\begin{equation*}
\begin{split}
    \| \x_{\hat{\lambda}_{\l}, \l} - \x_{\l} \|
    &\coloneqq \left( \sum_{i = 1}^{n_{\l}} | x_{\hat{\lambda}_{\l}, \l, i} - x_{\l, i} |^{2} \right)^{\frac{1}{2}} \\
    &\leq \sqrt{n_{\l}} \max_{i \in \{ 1, 2, \dots, n_{\l} \}} \{ | x_{\hat{\lambda}_{\l}, \l, i} - x_{\l, i} | \} \,,
\end{split}
\end{equation*}
where $x_{\hat{\lambda}_{\l}, \l, i}$ and $x_{\l, i}$ denote the $i$-th components of the $\l$-th layer regularised and quantised features, respectively.
We define
\begin{equation*}
    \bar{i} \coloneqq \mathop{\argmax}_{i \in \{ 1, 2, \dots, n_{\l} \}} \{ | x_{\hat{\lambda}_{\l}, \l, i} - x_{\l, i} | \} \,.
\end{equation*}
Therefore, since $n_{\l}$ is arbitrary but finite, a sufficient condition for \eqref{eq:convergence_thesis} is
\begin{equation}\label{eq:convergence_thesis_onedim}
    \frac{| x_{\hat{\lambda}_{\l}, \l, \bar{i}} - x_{\l, \bar{i}} |}{r^{\l}(\lambda)} \xrightarrow[\lambda \to 0]{} 0 \,.
\end{equation}
To simplify the notation, in the following we will omit the subscript index $\bar{i}$.
First, we conveniently rewrite \eqref{eq:convergence_thesis_onedim} according to the definition of limit:
\begin{equation}\label{eq:convergence_thesis_onedim_limdef}
\begin{split}
    &\forall\, \eps > 0 \,,\, \,\exists\, \tilde{\lambda} > 0 \,:\, \\
    &\qquad | x_{\hat{\lambda}_{\l}, \l} - x_{\l} | < \eps r_{\l}(\lambda) \,,\, \,\forall\, 0 < \lambda < \tilde{\lambda} \,.
\end{split}
\end{equation}
Then, we argue by induction.

To prove the base step ($\l = 1$) we need to consider two cases: $x_{1} = 0$ and $x_{1} = 1$.
First, we suppose $x_{1} = \sigma(S_{\m_{1}}(\x_{0})) = 0$; this implies that $S_{\m_{1}}(\x_{0}) < 0$.
Property~\eqref{eq:regparam_hp_iv} implies $x_{\hat{\lambda}_{1}, 1} \geq x_{1}$, which implies $| x_{\hat{\lambda}_{1}, 1} - x_{1} | = x_{\hat{\lambda}_{1}, 1} = \sigma_{\lambda_{1}}(S_{\m_{1}}(\x_{0}))$.
Then, we can apply $\sigma_{\lambda_{1}}^{-1}$ to both sides of the inequality in \eqref{eq:convergence_thesis_onedim_limdef} obtaining the following condition:
\begin{equation*}
\begin{split}
    &\forall\, \eps > 0 \,,\, \,\exists\, \tilde{\lambda} > 0 \,:\, \\
    &\qquad S_{\m^{1}}(\x^{0}) < \sigma_{\lambda_{1}}^{-1}(\eps r_{1}(\lambda)) \,,\, \,\forall\, 0 < \lambda < \tilde{\lambda} \,,
\end{split}
\end{equation*}
whose validity is guaranteed by hypothesis \eqref{eq:convergence_hp_i}.
Now, we analyse the case $x_{1} = 1$.
Property~\eqref{eq:regparam_hp_iv} implies $x_{\hat{\lambda}_{1}, 1} \leq x_{1}$, hence $| x_{\hat{\lambda}_{1}, 1} - x_{1} | = 1 - x_{\hat{\lambda}_{1}, 1} = 1 - \sigma_{\lambda_{1}}(S_{\m_{1}}(\x_{0}))$.
In this case, condition \eqref{eq:convergence_thesis_onedim_limdef} becomes
\begin{equation}\label{eq:convergence_thesis_onedim_limdef_transf}
\begin{split}
    &\forall\, \eps > 0 \,,\, \,\exists\, \tilde{\lambda} > 0 \,:\, \\
    &\qquad 1 - \sigma_{\lambda_{1}}(S_{\m_{1}}(\x_{0})) < \eps r_{1}(\lambda) \,,\, \,\forall\, 0 < \lambda < \tilde{\lambda} \,.
\end{split}
\end{equation}
We have two sub-cases: $S_{\m_{1}}(\x_{0}) > 0$ and $S_{\m_{1}}(\x_{0}) = 0$.
In the first sub-case, by rearranging terms and applying $\sigma_{\lambda_{1}}^{-1}$ to both sides of \eqref{eq:convergence_thesis_onedim_limdef_transf}, we derive the condition
\begin{equation*}
\begin{split}
    &\forall\, \eps > 0 \,,\, \,\exists\, \tilde{\lambda} > 0 \,:\, \\
    &\qquad \sigma_{\lambda_{1}}^{-1}(1 - \eps r_{1}(\lambda)) < S_{\m_{1}}(\x_{0}) \,,\, \,\forall\, 0 < \lambda < \tilde{\lambda} \,,
\end{split}
\end{equation*}
which is granted by hypothesis \eqref{eq:convergence_hp_ii}.
In the second sub-case, we can divide both sides of \eqref{eq:convergence_thesis_onedim_limdef_transf} by $r_{1}(\lambda)$ and obtain the condition
\begin{equation*}
\begin{split}
    &\forall\, \eps > 0 \,,\, \,\exists\, \tilde{\lambda} > 0 \,:\, \\
    &\qquad \frac{1 - \sigma_{\lambda_{1}}(0)}{r_{1}(\lambda)} < \eps \,,\, \,\forall\, 0 < \lambda < \tilde{\lambda} \,,
\end{split}
\end{equation*}
which holds by hypothesis \eqref{eq:convergence_hp_iii}.

We now proceed to the inductive step ($\l > 1$).
We have two possibilities for $x_{\l}$:
\begin{enumerate}[label=(\Alph*)]
    \item\label{th:convergence_case_A} $x_{\l} = 0$;
    \item\label{th:convergence_case_B} $x_{\l} = 1$.
\end{enumerate}
We start with case \ref{th:convergence_case_A}.
We observe that
\begin{equation}\label{eq:convergence_caseA}
\begin{split}
    s_{\hat{\lambda}_{\l-1}, \l} - s_{\l}
    &= S_{\m_{\l}}(\x_{\hat{\lambda}_{\l-1}, \l-1}) - S_{\m_{\l}}(\x_{\l-1}) \\
    &= S_{\m_{\l}}(\x_{\hat{\lambda}_{\l-1}, \l-1} - \x_{\l-1}) 
    \xrightarrow[\lambda \to 0]{} 0 \,,
\end{split}
\end{equation}
since $S_{\m_{\l}}$ is linear and $\| \x_{\hat{\lambda}_{\l-1}, \l-1} - \x_{\l-1} \| \xrightarrow[\lambda \to 0]{} 0$ by the inductive hypothesis.
With reference to $H_{0}^{+}$, case \ref{th:convergence_case_A} implies that $s_{\l} < 0$.
Together with \eqref{eq:convergence_caseA}, this implies that
\begin{equation*}
\begin{split}
    &\exists\, \lambda^{*} = \lambda^{*}(s_{\l}) > 0 \,:\, \\
    &\qquad s_{\hat{\lambda}_{\l-1}, \l} < -\frac{|s_{\l}|}{2} < 0 \,,\, \,\forall\, 0 < \lambda < \lambda^{*} \,.
\end{split}
\end{equation*}
Since $x_{\l} = 0$ and $x_{\hat{\lambda}_{\l}, \l} = \sigma_{\lambda_{\l}}(s_{\hat{\lambda}_{\l-1}, \l}) \geq 0$, condition \eqref{eq:convergence_thesis_onedim_limdef} can be rewritten as
\begin{equation*}
\begin{split}
    &\forall\, \eps > 0 \,,\, \,\exists\, \tilde{\lambda} > 0 \,:\, \\
    &\qquad \sigma_{\lambda_{\l}}(s_{\hat{\lambda}_{\l-1}, \l}) < \eps r_{\l}(\lambda) \,,\, \,\forall\, 0 < \lambda < \tilde{\lambda} \,.
\end{split}
\end{equation*}
Due to the monotonicity of $\sigma_{\lambda_{\l}}$, we have $\sigma_{\lambda_{\l}}(s_{\hat{\lambda}_{\l-1}, \l}) < \sigma_{\lambda_{\l}}(-|s_{\l}|/2) \,,\, \,\forall\, 0 < \lambda < \lambda^{*}$.
Therefore, a sufficient condition to guarantee the convergence is that
\begin{equation*}
\begin{split}
    &\forall\, \eps > 0 \,,\, \,\exists\, 0 < \tilde{\lambda} \leq \lambda^{*} \,:\, \\
    &\qquad -\frac{|s_{\l}|}{2} < \sigma_{\lambda_{\l}}^{-1}(\eps r_{\l}(\lambda)) \,,\, \,\forall\, 0 < \lambda < \tilde{\lambda} \,.
\end{split}
\end{equation*}
This condition is granted for every $s_{\l} < 0$ by \eqref{eq:convergence_hp_i}.
We now move to case \ref{th:convergence_case_B}.
This case ($x_{\l} = 1$) might originate from two sub-cases:
\begin{enumerate}[label=(\roman*)]
    \item\label{th:convergence_case_Bi} $s_{\l} > 0$;
    \item\label{th:convergence_case_Bii} $s_{\l} = 0$.
\end{enumerate}
The proof of sub-case \ref{th:convergence_case_Bi} is similar to the proof for case \ref{th:convergence_case_A}.
Given $s_{\l} > 0$, since $s_{\hat{\lambda}_{\l-1}, \l} \xrightarrow[\lambda \to 0]{} s_{\l}$ by the inductive hypothesis, we have that
\begin{equation*}
\begin{split}
    &\exists\, \lambda^{*} = \lambda^{*}(s_{\l}) > 0 \,:\, \\
    &\qquad 0 < \frac{s_{\l}}{2} < s_{\hat{\lambda}_{\l-1}, \l} \,.
\end{split}
\end{equation*}
Then, since $x_{\l} = 1$ and $x_{\l} = \sigma_{\lambda_{\l}}(s_{\hat{\lambda}_{\l-1}, \l}) \leq 1$, we can rewrite \eqref{eq:convergence_thesis_onedim_limdef} as
\begin{equation*}
\begin{split}
    &\forall\, \eps > 0 \,,\, \,\exists\, \tilde{\lambda} > 0 \,:\, \\
    &\qquad 1 - \sigma_{\lambda_{\l}}(s_{\hat{\lambda}_{\l-1}, \l}) < \eps r_{\l}(\lambda) \,,\, \,\forall\, 0 < \lambda < \tilde{\lambda} \,.
\end{split}
\end{equation*}
Since $\sigma_{\lambda_{\l}}(s_{\hat{\lambda}_{\l-1}, \l}) > \sigma_{\lambda_{\l}}(s_{\l}/2)$, a sufficient condition to get convergence is that
\begin{equation*}
\begin{split}
    &\forall\, \eps > 0 \,,\, \,\exists\, 0 < \tilde{\lambda} < \lambda^{*} \,:\, \\
    &\qquad \frac{s_{\l}}{2} > \sigma_{\lambda_{\l}}^{-1}(1 - \eps r_{\l}(\lambda)) \,,\, \,\forall\, 0 < \lambda < \tilde{\lambda} \,.
\end{split}
\end{equation*}
This is guaranteed for every $s_{\l} > 0$ by \eqref{eq:convergence_hp_i}.
Case \ref{th:convergence_case_Bii} is more delicate, since $s_{\hat{\lambda}_{\l-1}, \l}$ can be positioned in two ways with respect to $s_{\l} = 0$:
\begin{enumerate}[label=(\alph*)]
    \item\label{th:convergence_case_Biia} $\lambda > 0$ is such that $s_{\hat{\lambda}_{\l-1}, \l} \geq 0$;
    \item\label{th:convergence_case_Biib} $\lambda > 0$ is such that $s_{\hat{\lambda}_{\l-1}, \l} < 0$.
\end{enumerate}
In case \ref{th:convergence_case_Biia}, it is sufficient to note that the monotonicity of $\sigma_{\lambda_{\l}}$ implies $1 - \sigma_{\lambda_{\l}}(s_{\hat{\lambda}_{\l-1}, \l}) \leq 1 - \sigma_{\lambda_{\l}}(0)$, since $s_{\hat{\lambda}_{\l-1}, \l} \geq 0$.
Then, condition \eqref{eq:convergence_thesis_onedim_limdef} can be rewritten as
\begin{equation*}
\begin{split}
    &\forall\, \eps > 0 \,,\, \,\exists\, \tilde{\lambda} > 0 \,:\, \\
    &\qquad \frac{1 - \sigma_{\lambda_{\l}}(0)}{r_{\l}(\lambda)} < \eps \,,\, \,\forall\, 0 < \lambda < \tilde{\lambda} \,.
\end{split}
\end{equation*}
This is guaranteed by \eqref{eq:convergence_hp_iii}.
To prove the last case \ref{th:convergence_case_Biib}, we first observe that
\begin{equation*}
\begin{split}
    s_{\hat{\lambda}_{\l-1}, \l}
    &= s_{\hat{\lambda}_{\l-1}, \l} - s_{\l} \\
    &= \left( \langle \w_{\l}, \x_{\hat{\lambda}_{\l-1}, \l-1} \rangle + b_{\l} \right) - \left( \langle \w_{\l}, \x_{\l-1} \rangle + b_{\l} \right) \\
    &= \langle \w_{\l}, \x_{\hat{\lambda}_{\l-1}, \l-1} - \x_{\l-1} \rangle
\end{split}
\end{equation*}
(which follows from $s_{\l} = 0$) and apply the Cauchy-Schwartz inequality to obtain the following upper bound:
\begin{equation}\label{eq:convergence_cauchyschwartz}
    |s_{\hat{\lambda}_{\l-1}, \l}| \leq \| \w_{\l} \| \, \| \x_{\hat{\lambda}_{\l-1}, \l-1} - \x_{\l-1} \| \,.
\end{equation}
Then, we rewrite \eqref{eq:convergence_thesis_onedim_limdef} as
\begin{equation*}
\begin{split}
    &\forall\, \eps > 0 \,,\, \,\exists\, \tilde{\lambda} > 0 \,:\, \\
    &\qquad -s_{\hat{\lambda}_{\l-1}, \l} < -\sigma_{\lambda_{\l}}^{-1}(1 - \eps r_{\l}(\lambda)) \,,\, \,\forall\, 0 < \lambda < \tilde{\lambda} \,.
\end{split}
\end{equation*}
Observation \eqref{eq:convergence_cauchyschwartz} allows us to write a slightly stronger but sufficient condition for convergence:
\begin{equation*}
\begin{split}
    &\forall\, \eps > 0 \,,\, \,\exists\, \tilde{\lambda} > 0 \,:\, \\
    &\qquad \| \w_{\l} \| \, \| \x_{\hat{\lambda}_{\l-1}, \l-1} - \x_{\l-1} \| \leq -\sigma_{\lambda_{\l}}^{-1}(1 - \eps r_{\l}(\lambda)) \,, \\
    &\qquad \forall\, 0 < \lambda < \tilde{\lambda} \,,
\end{split}
\end{equation*}
where we used the fact that $-s_{\hat{\lambda}_{\l-1}, \l} = | s_{\hat{\lambda}_{\l-1}, \l} |$ (since $s_{\hat{\lambda}_{\l-1}, \l} < 0$).
The inner inequality can be rewritten as
\begin{equation}\label{eq:convergence_caseBiib_transf}
    \frac{\| \x_{\hat{\lambda}_{\l-1}, \l-1} - \x_{\l-1} \|}{-\sigma_{\lambda_{\l}}^{-1}(1 - \eps r_{\l}(\lambda))} \leq \frac{1}{\| \w_{\l}\|} \,;
\end{equation}
since $\w_{\l}$ is fixed but arbitrary (it is part of the parameter $\m_{\l}$), the term on the right can be arbitrarily small, and therefore a sufficient condition to ensure \eqref{eq:convergence_caseBiib_transf} for $\lambda$ small enough is
\begin{equation}\label{eq:convergence_caseBiib_suffcond}
    \forall\, \eps > 0 \,,\, \frac{\| \x_{\hat{\lambda}_{\l-1}, \l-1} - \x_{\l-1} \|}{-\sigma_{\lambda_{\l}}^{-1}(1 - \eps r_{\l}(\lambda))} \xrightarrow[\lambda \to 0]{} 0 \,.
\end{equation}
By the inductive hypothesis (where we set $\eps = 1$),
\begin{equation*}
\begin{split}
    &\exists\, \tilde{\lambda}_{\l-1} > 0 \,:\, \\
    &\qquad \| \x_{\hat{\lambda}_{\l-1}, \l-1} - \x_{\l-1} \| < r_{\l-1}(\lambda) \,,\, \,\forall\, 0 < \lambda < \tilde{\lambda}_{\l-1} \,.
\end{split}
\end{equation*}
Therefore, \eqref{eq:convergence_hp_iv} enforces the convergence of the upper bound in the following inequality:
\begin{equation*}
    \frac{\| \x_{\hat{\lambda}_{\l-1}, \l-1} - \x_{\l-1} \|}{-\sigma_{\lambda_{\l}}^{-1}(1 - \eps r_{\l}(\lambda))} \leq \frac{r_{\l-1}(\lambda)}{-\sigma_{\lambda_{\l}}^{-1}(1 - \eps r_{\l}(\lambda))} \,,
\end{equation*}
and therefore \eqref{eq:convergence_caseBiib_suffcond} follows.
This completes the proof of the theorem.
\end{proof}

\clearpage
\section{Other experiments}
\label{app:experiments}

The experimental findings reported in Section~\ref{sec:experiments} refer to a ternary VGG-like network solving CIFAR-10.
To corroborate the validity of our findings, we performed additional experiments on two different scenarios, where we changed the data set, the network topology, and the quantisation policy.

Given the findings reported in Section~\ref{sec:experiments}, we used different noise types only when analysing static noise schedules.
We constrained the noise type to uniform when analysing dynamic noise schedules.

As in the original CIFAR-10 experiments, we evaluated each hyper-parameter configuration using five-fold cross-validation on the training partitions of the chosen data sets.

\subsection{SVHN}

\gls{SVHN} is an image classification data set \cite{Netzer2011}.
It contains $\sim 99k$ RGB-encoded images representing decimal digits from house number plates.
It consists of a training partition ($\sim 73k$ images) and a validation partition ($\sim 26k$ images).

We used the same VGG-like network from the CIFAR-10 experiments.
Again, we quantised all the weights and features to be ternary, and we kept the weights of the last layer in floating-point format.

In each experimental unit, we trained the network for $500$ epochs using mini-batches of $256$ images, the cross-entropy loss function, and the ADAM optimiser with an initial learning rate of $10^{-3}$, decreased to $10^{-4}$ after $400$ epochs.

\begin{figure}[t]
	\begin{subfigure}[b]{1.0\linewidth}
	    \centering\includegraphics[width=1.0\linewidth]{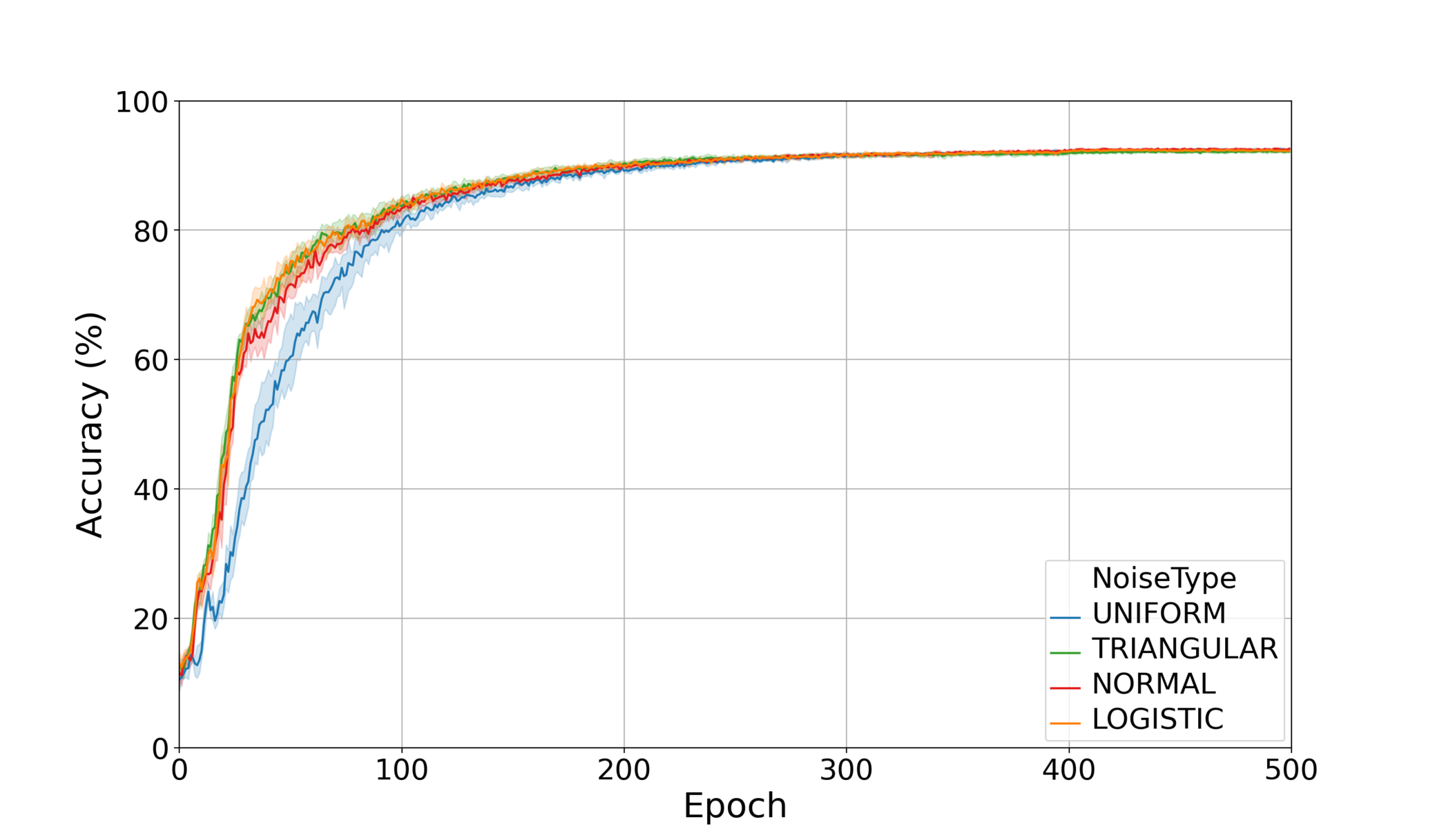}
    	\caption{}    \label{fig:svhn_static_schedule_random}
	\end{subfigure}
	\begin{subfigure}[b]{1.0\linewidth}
    	\centering\includegraphics[width=1.0\linewidth]{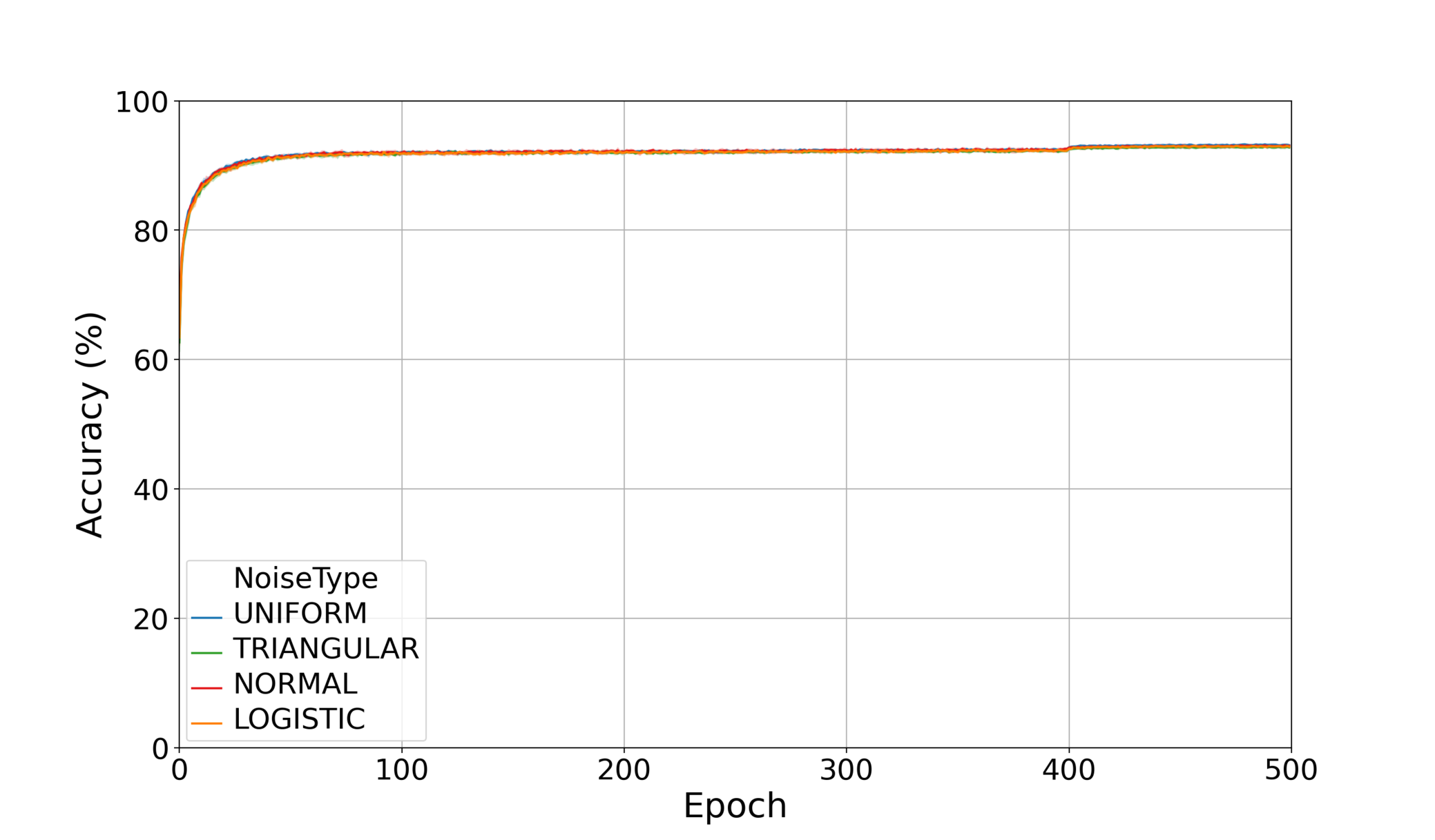}
	    \caption{}    \label{fig:svhn_static_schedule_mode}
	\end{subfigure}
	\caption{Performance of \gls{ANA} on the \gls{SVHN} data set using static noise schedules in combination with different forward computation strategies: random~\ref{fig:svhn_static_schedule_random}, mode~\ref{fig:svhn_static_schedule_mode}. Each plot reports different noise types using different colours: uniform (blue), triangular (green), normal (red), logistic (yellow).}
\end{figure}

In agreement with the CIFAR-10 findings, Figures~\ref{fig:svhn_static_schedule_random}~\ref{fig:svhn_static_schedule_mode} show that \glspl{QNN} trained using static \gls{STE} variants based on different noise types converge to the same accuracy.
We note that the uniform noise type in combination with the random forward computation strategy seems to perform slightly worse during the earlier stages of training.

\begin{figure}
	\begin{subfigure}[b]{1.0\linewidth}
	    \centering\includegraphics[width=1.0\linewidth]{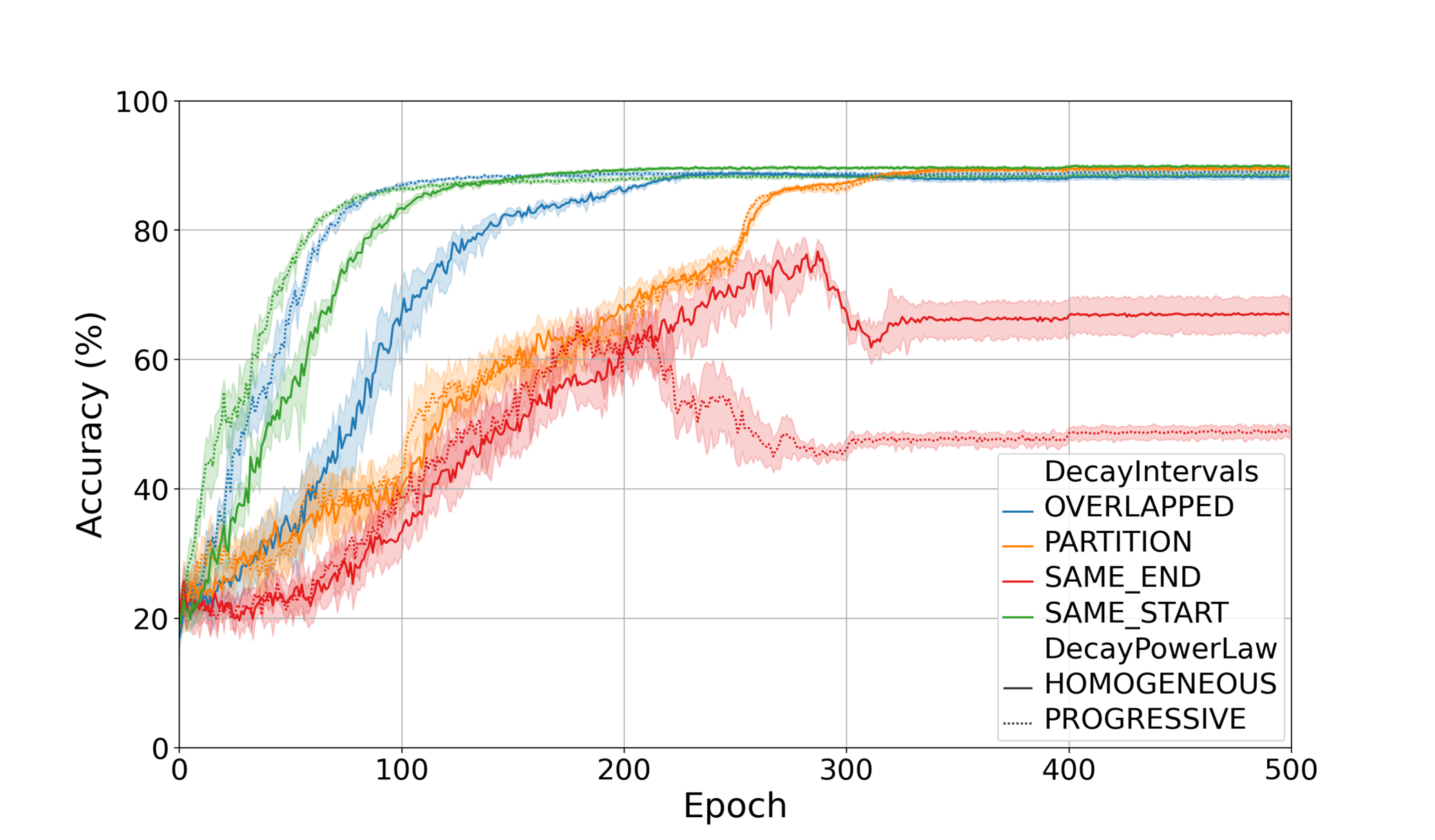}
    	\caption{}    \label{fig:svhn_uniform_dynamic_schedule_expectation}
	\end{subfigure}
	\begin{subfigure}[b]{1.0\linewidth}
	    \centering\includegraphics[width=1.0\linewidth]{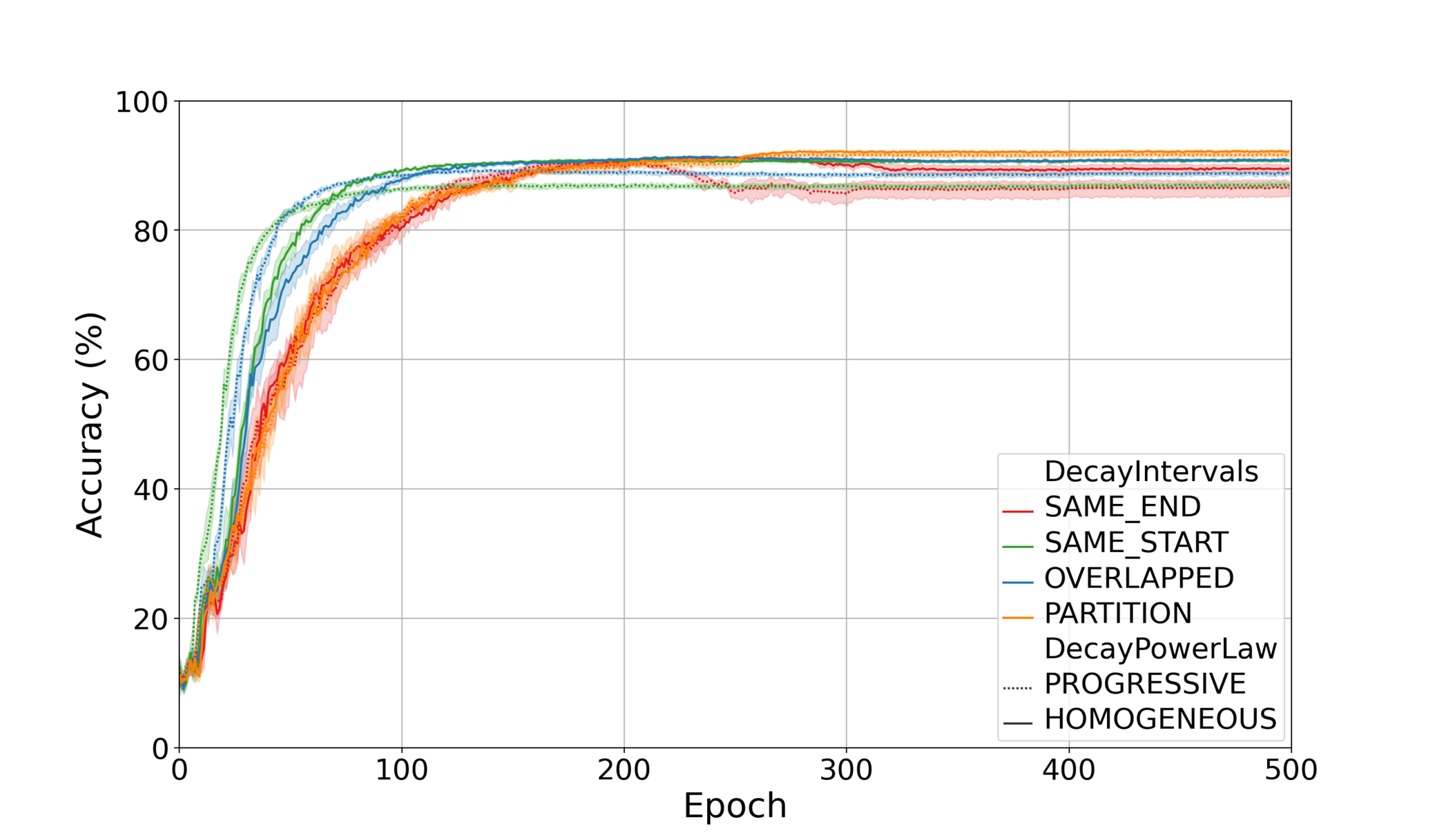}
    	\caption{}    \label{fig:svhn_uniform_dynamic_schedule_random}
	\end{subfigure}
	\begin{subfigure}[b]{1.0\linewidth}
	    \centering\includegraphics[width=1.0\linewidth]{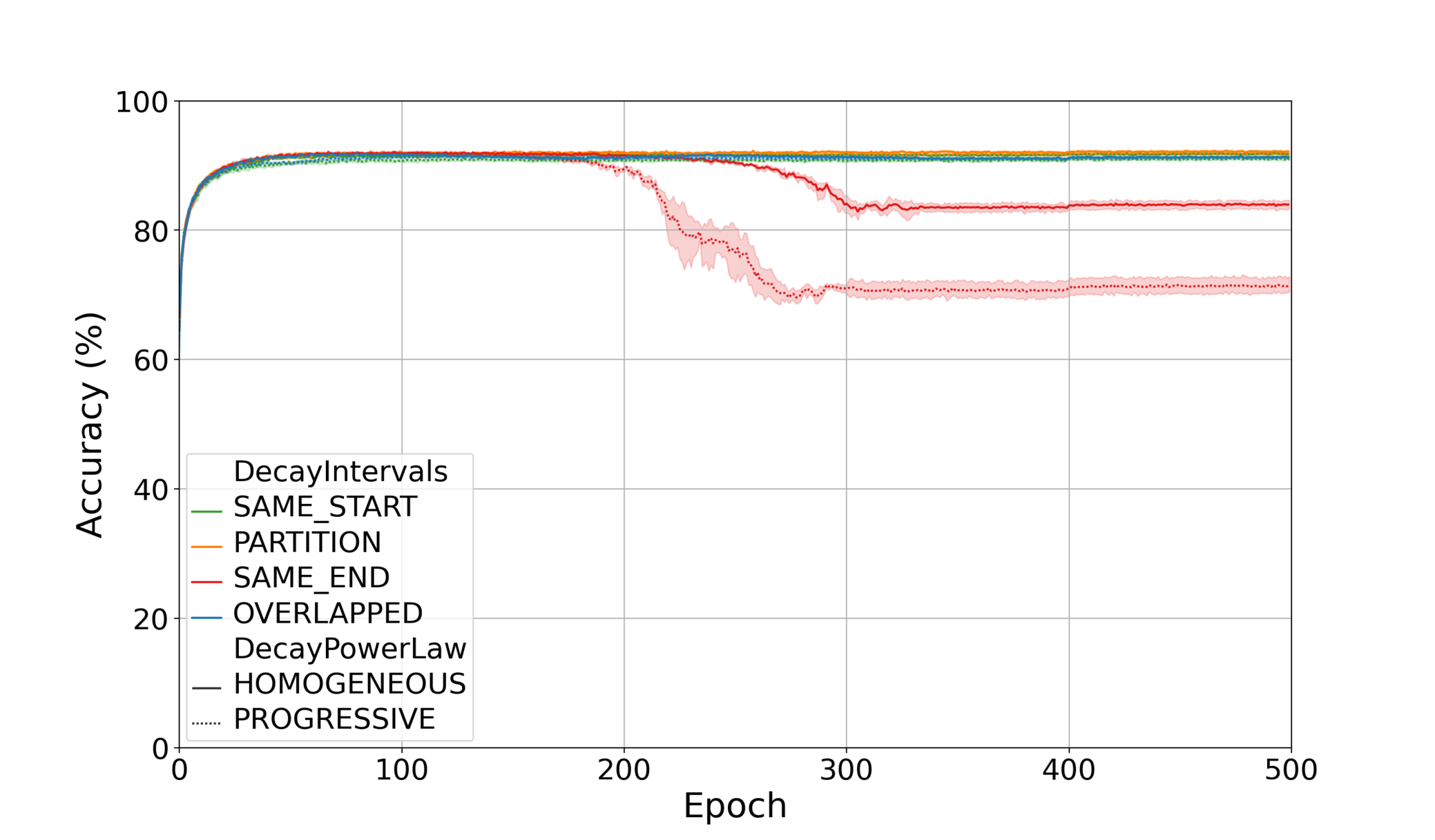}
    	\caption{}    \label{fig:svhn_uniform_dynamic_schedule_mode}
	\end{subfigure}
	\caption{Performance of \gls{ANA} on the \gls{SVHN} data set using dynamic noise schedules (static means, dynamic variances) under uniform noise and different forward computation strategies: expectation~\ref{fig:svhn_uniform_dynamic_schedule_expectation}, random~\ref{fig:svhn_uniform_dynamic_schedule_random}, mode~\ref{fig:svhn_uniform_dynamic_schedule_mode}. Each plot reports multiple schedules: decay intervals: same start (green), same end (red), partition (yellow), overlapped (blue);  decay power law: homogeneous (continuous), progressive (dotted).}
\end{figure}

Figures~\ref{fig:svhn_uniform_dynamic_schedule_expectation},~\ref{fig:svhn_uniform_dynamic_schedule_random},~\ref{fig:svhn_uniform_dynamic_schedule_mode} show that the quality of different decay interval strategies (as measured by the final accuracy of the trained networks) is better for those that are more coherent with the hypothesis of Theorem~\ref{th:compositional_convergence}, namely the partition and same start strategies.
Independently of the forward computation strategy, the same end decay interval strategy is still the worst amongst the tested ones.

\subsection{GSC}

\gls{GSC} is a keyword spotting data set \cite{Warden2018}.
Keyword spotting requires mapping word utterances to the corresponding items in a given vocabulary.
It is an elementary though important speech recognition task, having widespread applications to speech-based user interactions with embedded devices such as smartphones or smartwatches.
\gls{GSC} contains $\sim 106k$ one-second utterances of $35$ different keywords recorded at $16kHz$, plus recordings of random background noise.
There are different keyword spotting tasks associated with \gls{GSC}; in our experiments, we focussed on the simplified $12$-class classification problem.

We used the DSCNN network topology \cite{Zhang2017}, a fully-feedforward network topology consisting of eight convolutional layers (four blocks concatenating a depth-wise convolution with a point-wise one) and one fully-connected layer; therefore, $L = 9$.
This time, we quantised weights aiming for the \texttt{INT4} (signed) data type, and features aiming for the \texttt{UINT4} (unsigned) data type.
Coherently with literature practice, we kept the last layer in floating-point format.

In each experimental unit, we trained the network for $120$ epochs using mini-batches of $256$ pre-processed utterances, the cross-entropy loss function, and the ADAM optimiser with an initial learning rate of $10^{-3}$, decreased to $10^{-4}$ after $100$ epochs.

\begin{figure}[t]
	\begin{subfigure}[b]{1.0\linewidth}
	    \centering\includegraphics[width=1.0\linewidth]{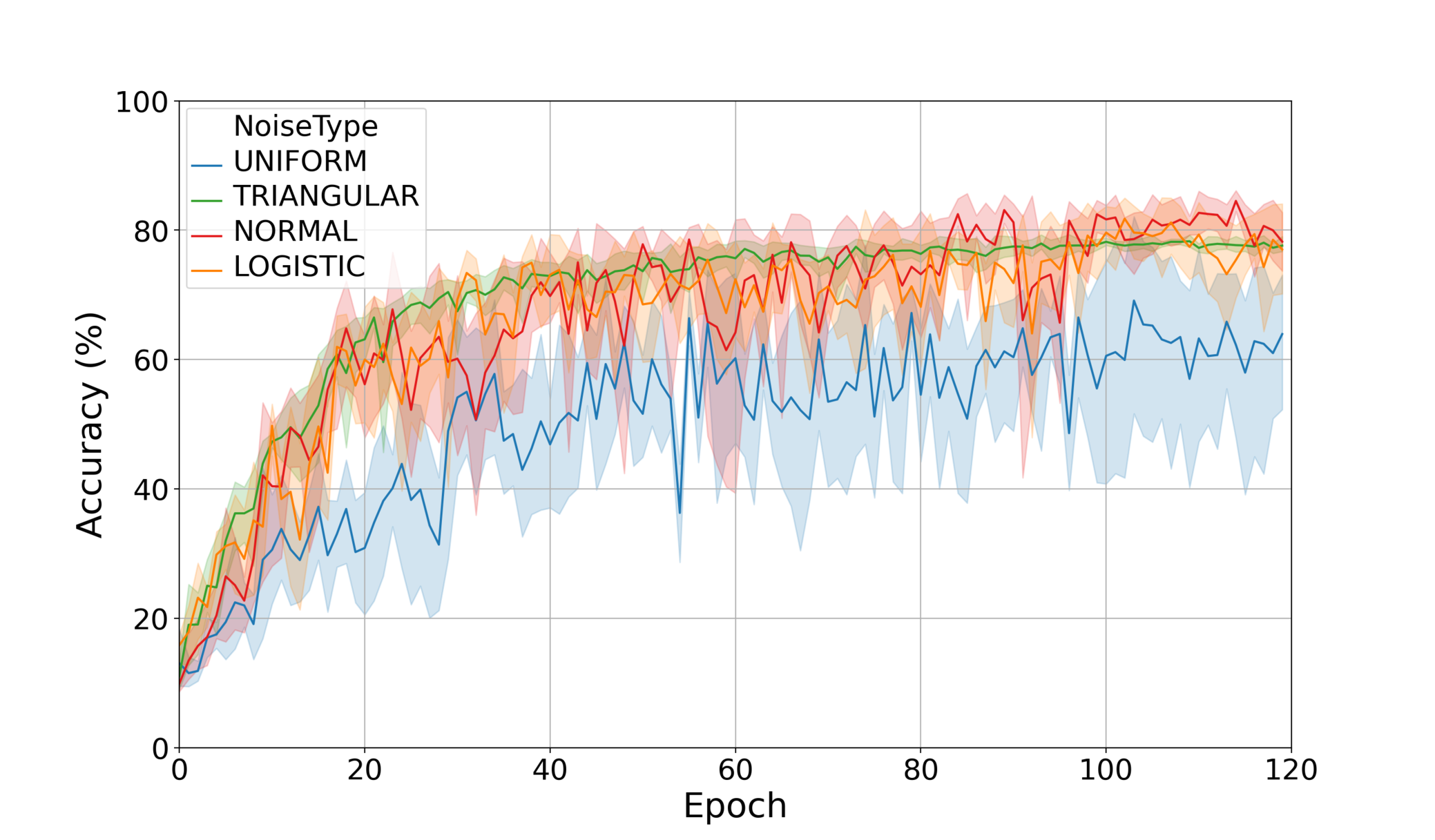}
    	\caption{}    \label{fig:gsc_static_schedule_random}
	\end{subfigure}
	\begin{subfigure}[b]{1.0\linewidth}
	    \centering\includegraphics[width=1.0\linewidth]{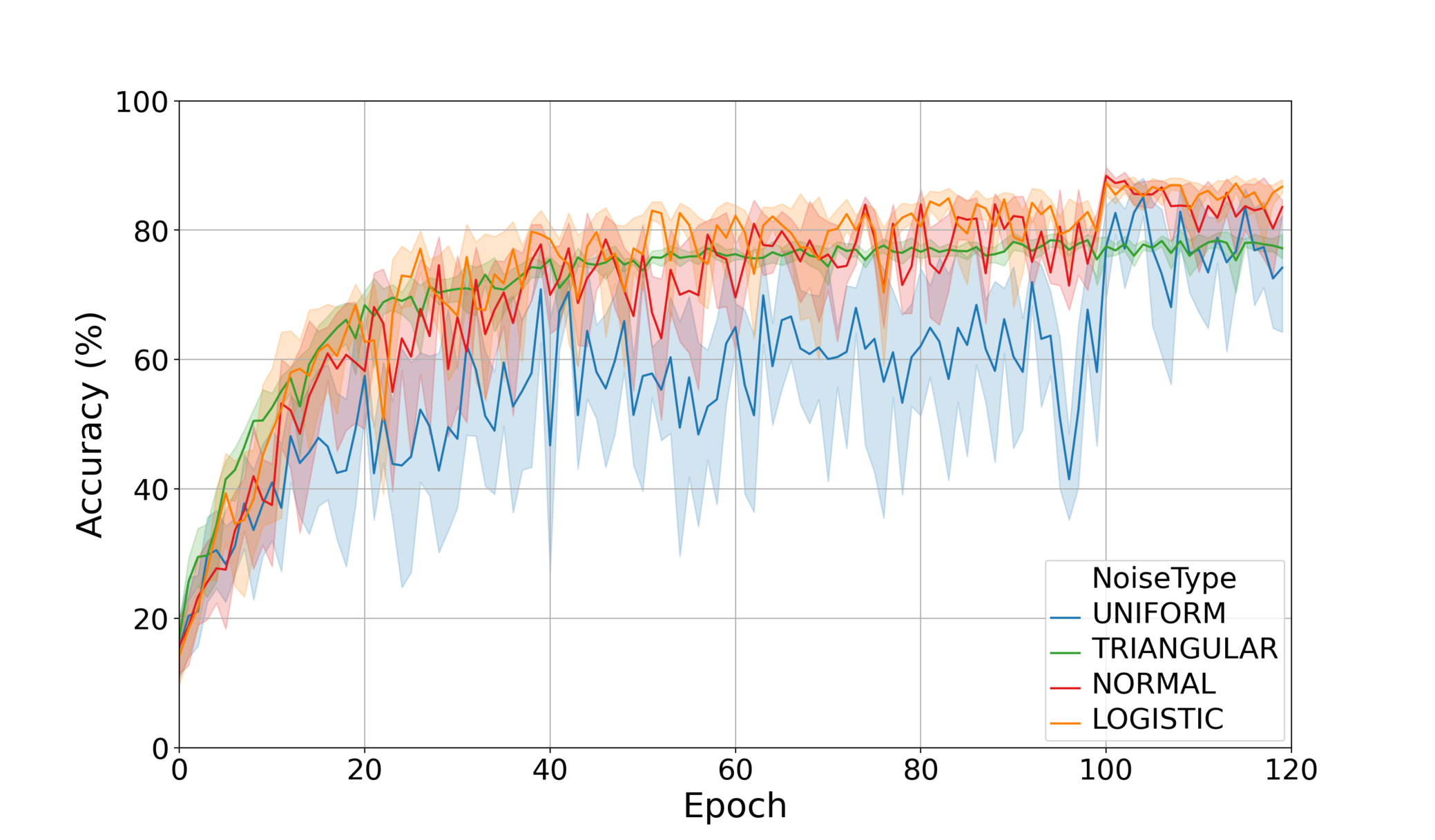}
    	\caption{}    \label{fig:gsc_static_schedule_mode}
	\end{subfigure}
	\caption{Performance of \gls{ANA} on the \gls{GSC} data set using static noise schedules in combination with different forward computation strategies: random~\ref{fig:gsc_static_schedule_random}, mode~\ref{fig:gsc_static_schedule_mode}. Each plot reports different noise types using different colours: uniform (blue), triangular (green), normal (red), logistic (yellow).}
\end{figure}

Figures~\ref{fig:gsc_static_schedule_random}~\ref{fig:gsc_static_schedule_mode} show that \glspl{QNN} trained using static \gls{STE} variants based on different noise types still converge to approximately the same accuracy.
However, in this scenario we can observe that the accuracy of \glspl{QNN} trained using the triangular noise type has lower variability, whereas that of \glspl{QNN} trained using uniform noise has higher variability.

\begin{figure}
	\begin{subfigure}[b]{1.0\linewidth}
	    \centering\includegraphics[width=1.0\linewidth]{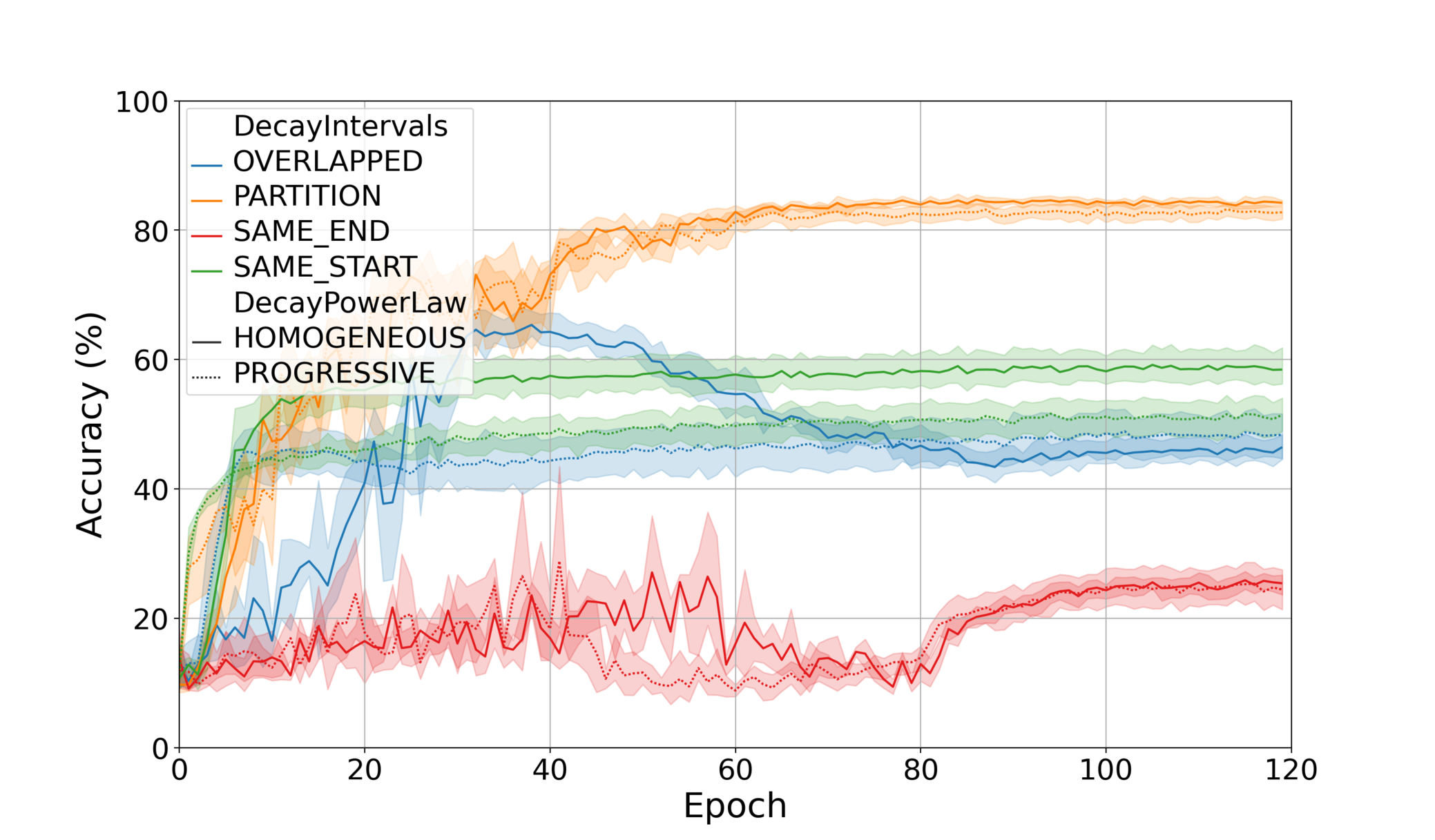}
    	\caption{}    \label{fig:gsc_uniform_dynamic_schedule_expectation}
	\end{subfigure}
	\begin{subfigure}[b]{1.0\linewidth}
	    \centering\includegraphics[width=1.0\linewidth]{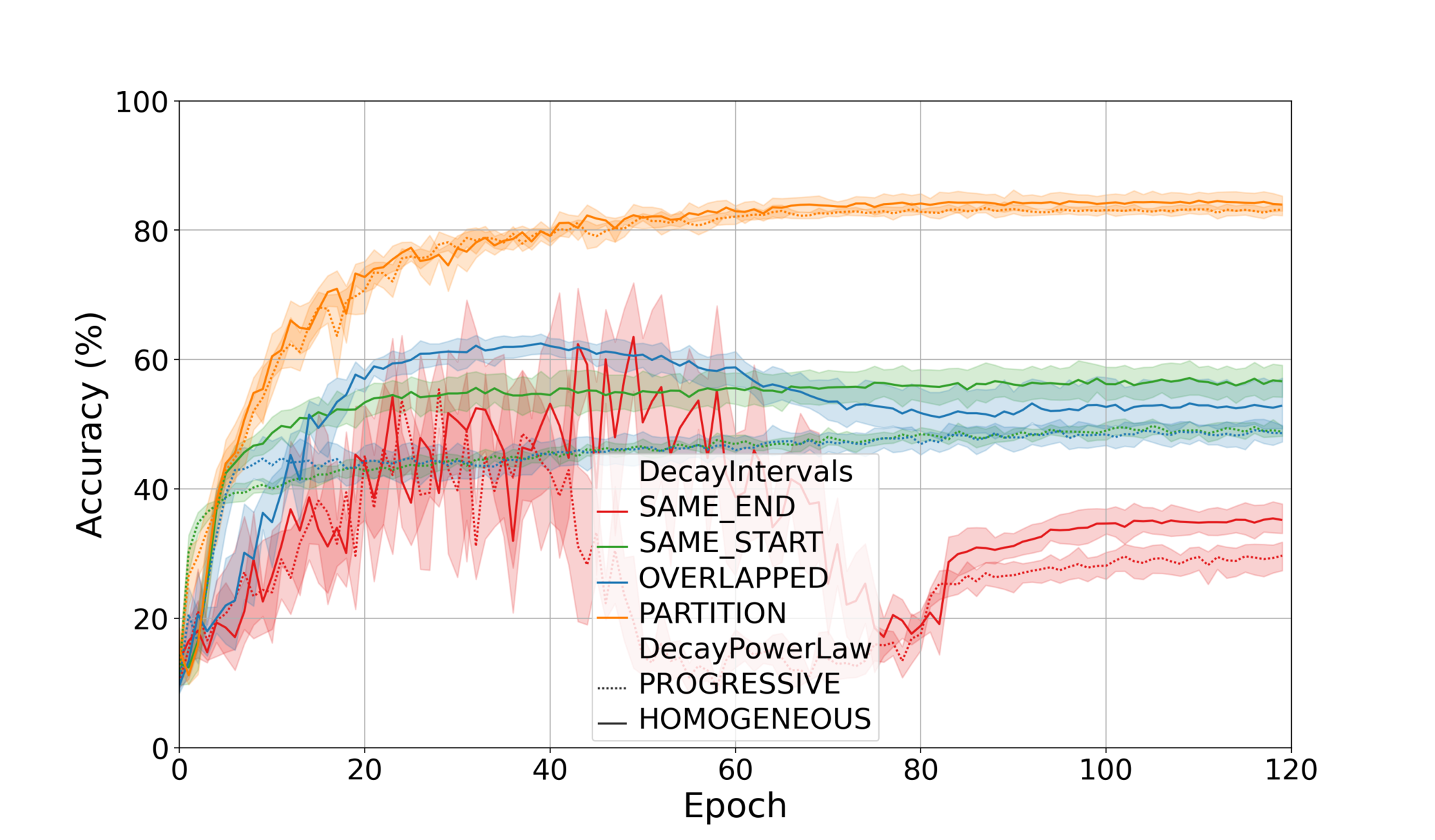}
    	\caption{}    \label{fig:gsc_uniform_dynamic_schedule_random}
	\end{subfigure}
	\begin{subfigure}[b]{1.0\linewidth}
	    \centering\includegraphics[width=1.0\linewidth]{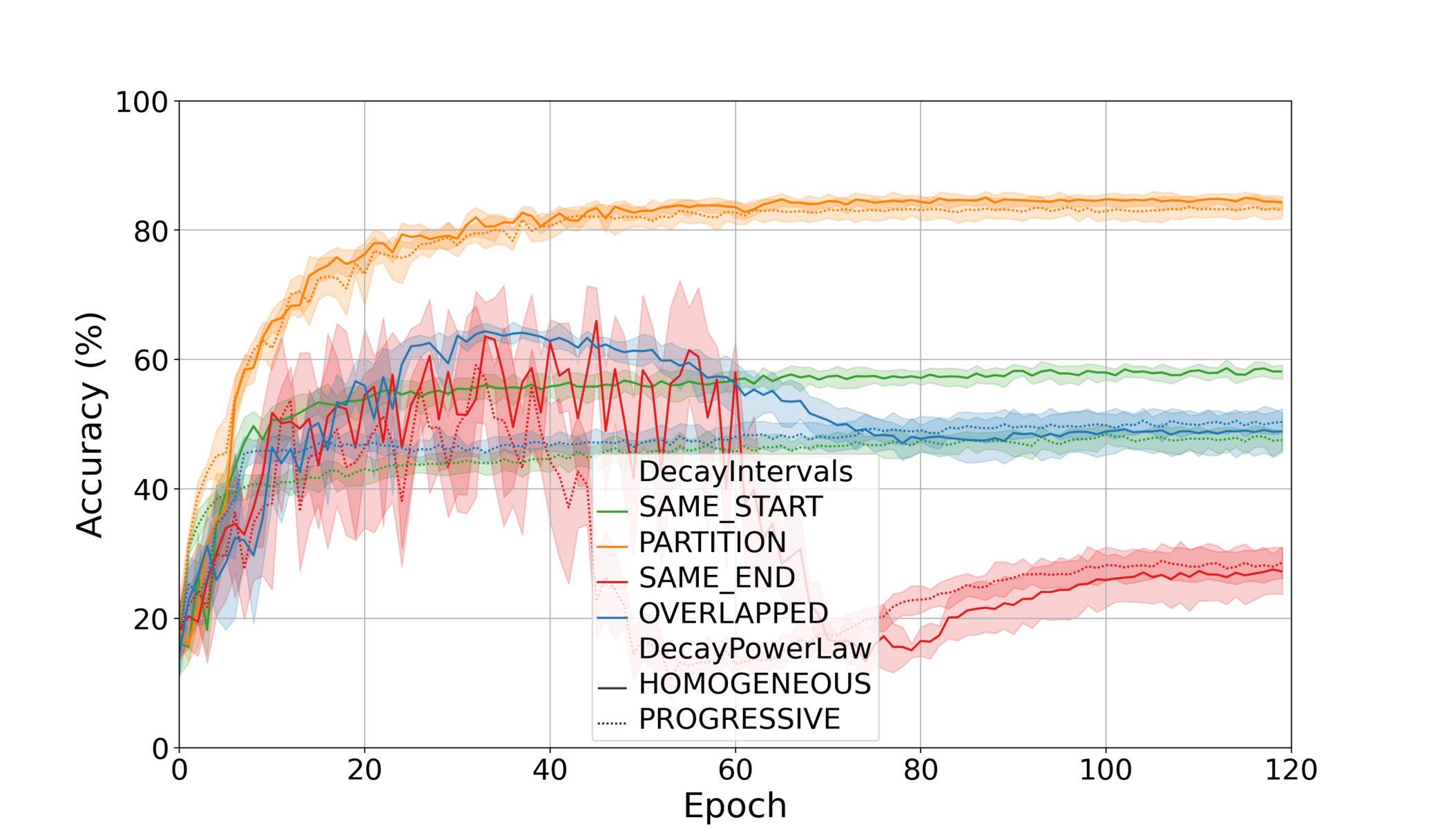}
    	\caption{}    \label{fig:gsc_uniform_dynamic_schedule_mode}
	\end{subfigure}
	\caption{Performance of \gls{ANA} on the \gls{GSC} data set using dynamic noise schedules (static means, dynamic variances) under uniform noise and different forward computation strategies: expectation~\ref{fig:gsc_uniform_dynamic_schedule_expectation}, random~\ref{fig:gsc_uniform_dynamic_schedule_random}, mode~\ref{fig:gsc_uniform_dynamic_schedule_mode}. Each plot reports multiple schedules: decay intervals: same start (green), same end (red), partition (yellow), overlapped (blue);  decay power law: homogeneous (continuous), progressive (dotted).}
\end{figure}

Figures~\ref{fig:gsc_uniform_dynamic_schedule_expectation},~\ref{fig:gsc_uniform_dynamic_schedule_random},~\ref{fig:gsc_uniform_dynamic_schedule_mode} show that the quality of different decay interval strategies (as measured by the final accuracy of the trained networks) is better for those that are more coherent with the hypothesis of Theorem~\ref{th:compositional_convergence}.
In particular, \glspl{QNN} trained using the partition strategy can achieve approximately the same accuracy as networks trained using static noise schedules.
Independently of the forward computation strategy, the same end decay interval strategy is still the worst amongst the tested ones.

\end{document}